\documentclass[journal]{IEEEtran}
\ifCLASSINFOpdf
\else
\fi
\usepackage{subfigure}
\usepackage{amssymb} 
\usepackage{amsmath} 
\usepackage{amsthm}
\usepackage{cases}
\usepackage{algorithmic}
\usepackage{algorithm}
\usepackage{graphicx}
\usepackage{bm}
\usepackage{color}
\usepackage{psfrag} 
\usepackage[process=auto]{pstool}
\usepackage{threeparttable}
\usepackage{dcolumn}
\usepackage{multirow} 
\usepackage{float}
\usepackage{setspace}
\usepackage{cite}
\usepackage{amsfonts}
\usepackage{textcomp}
\usepackage{epsfig} 
\usepackage{mathptmx} 
\usepackage{times} 
\usepackage{bookmark}
\usepackage{ifpdf}
\usepackage{scalerel}
\usepackage{makecell}
\usepackage{stfloats}
\usepackage{booktabs}
\usepackage{placeins}
\usepackage[dvipsnames]{xcolor}
\usepackage[numbers,sort&compress]{natbib}
\begin{document}
%
\title{A Fast Planning Approach for 3D Short Trajectory with a Parallel Framework}
%
%
%
\author{Han Chen$^1$, Shengyang Chen$^2$, Peng Lu$^{3*}$, and Chih-Yung Wen$^2$ 
\thanks{$^1$Department of Aeronautical and Aviation Engineering, Hong Kong Polytechnic University, Hong Kong, China.
}
\thanks{$^2$Department of Mechanical Engineering, The Hong Kong Polytechnic University, Hong Kong, China.
}
\thanks{$^3$Department of Mechanical Engineering, The University of Hong Kong, Hong Kong, China.
{\tt\small lupeng@hku.hk}}
\thanks{$^*$Corresponding author}}

%
%

\markboth{IEEE transactions ON ROBOTICS}%
{Shell \MakeLowercase{\textit{et al.}}: Bare Demo of IEEEtran.cls for IEEE Journals}
%



\maketitle

\begin{abstract}
For real applications of unmanned aerial vehicles, the capability of navigating with full autonomy in unknown environments is a crucial requirement. However, planning a shorter path with less computing time is contradictory.
To address this problem, we present a framework with the map planner and point cloud planner running in parallel in this paper. The map planner determines the initial path using the improved jump point search method on the 2D map, and then it tries to optimize the path by considering a possible shorter 3D path.
The point cloud planner is executed at a high frequency to generate the motion primitives. It makes the drone follow the solved path and avoid the suddenly appearing obstacles nearby. Thus, vehicles can achieve a short trajectory while reacting quickly to the intruding obstacles.

We demonstrate fully autonomous quadrotor flight tests in unknown and complex environments with static and dynamic obstacles to validate the proposed method. In simulation and hardware experiments, the proposed framework shows satisfactorily comprehensive performance.
\end{abstract}


\maketitle
\section*{Supplementary Materials} \label{sec:supplementaryMaterial}
Demo video: \url{https://youtu.be/AOENvwf8sfM} \par

%
\IEEEpeerreviewmaketitle

\section{Introduction}
%
%
%
%
\IEEEPARstart{I}{n} application scenarios, such as searches and expeditions, small drones are usually used to explore unknown environments. To achieve autonomous unmanned aerial vehicle (UAV) navigation in unknown environments, achieving onboard simultaneous localization and mapping (SLAM) and path planning is required. In the research field of path planning for UAVs, the three most essential indicators are usually safety, path length, and calculation time for replanning the trajectory. In general, all methods are designed to ensure that the flight is safe, that is, collision-free. However, regarding path length and calculation time, most researchers have only focused on one of these factors because of the potential conflicts between them. In other words, calculating a shorter path is always more time-consuming. Minimizing the calculation time often requires direct planning based on the raw sensing data instead of planning on the periodically updated map, and therefore, it is difficult to handle complex environments. The UAV is more likely to detour, resulting in an inefficient flight trajectory. 

In addition, calculating the shortest path in the global 3D map consumes too much time and is not applicable to real-time planning. While flying in an unknown environment, the environmental information sensed by the drone is continuously used to update the map. After the map is updated, if the planned path cannot be replanned in time, the flight of the drone will be greatly imperiled. Therefore, for real-time calculations during drone flight, using a local map (the part of the global map around the location of the drone) is a common and effective method. Moreover, drones in unknown environments usually do not have a complete map, which means that the globally shortest path is difficult to plan.

Admittedly, planning the globally shortest path with only a local map is impossible. Shortening the path in the local map will also contribute to shortening the final flight path length. Nevertheless, to respond to emergencies, planning on the map may not be sufficiently fast. Mapping and planning on the map cost too much time to avoid the intruding dynamic obstacles. The drone must be able to avoid sudden obstacles in the unknown environment before the map is updated. Therefore, we propose a framework in which a low-frequency path planner and a high-frequency trajectory planner work in parallel. The designated goal is cast to the local map as the local goal. The map planner (MP) is first used to determine the 2D path to the local goal with the improved jump point search (JPS) method on the projection map. Then, a discrete angular graph search (DAGS) is used twice to find a 3D path that is obviously shorter than the 2D one. If the shorter 3D path is found, it is adopted. Otherwise, the 2D path is output. The point cloud planner (PCP) for trajectory planning is based on the design in our previous work \cite{chen2020computationally}. With a given goal, the PCP generates collision-free motion primitives continuously in a computationally efficient way to navigate the drone. In this parallel framework, it calculates the goal from the path output by the MP. In addition, we introduce the calculation formula for obtaining the goal for the PCP from the waypoints in the path. One benefit of the local map is that the computational time for the path planning on the map will not increase with the global map size. The path output frequency and the computational resource usage are guaranteed in a specific range to ensure that the loop frequency of the PCP is unaffected, and the MP can respond to the map change in time. In this framework, all the submodules are designed to minimize the time cost. For UAVs' real-time planning, it is safer when the planning outer loop frequency is higher.

The main contributions of this work are as follows:
\begin{itemize}
\item A parallel architecture with the MP and PCP is proposed, considering the planning success rate, path length, and fast response. The framework has been tested to achieve satisfactorily synthesized performance in extensive environments.

\item We improve the original JPS path and obtain a shorter 2D path. A sliding local map with two resolutions is introduced to increase the planning speed while maintaining a fine-grained path around the drone.

\item We introduce the DAGS based on the angular cost and try to find a 3D path shorter than the improved 2D path.

\item Based on our former work \cite{chen2020computationally}, the PCP is further optimized in time cost, motion planning success rate, and safety. To connect two planners in one framework, we build an optimization problem to calculate the local goal from the path output by the MP. The analytical solution of the optimization problem is found from a geometric view.
\end{itemize}

Our proposed framework's performance is tested and verified in several simulation and hardware tests. The flight trajectory length and the detailed algorithm execution time are compared with the shortest global path length and those of the state-of-the-art algorithms, respectively, demonstrating the superiority of our method with the best all-around performance.

 

\section{Related work}
\subsection{Environmental information retrieving method}
In the related works, two main categories for environmental information retrieving methods can be summarized: memory-less and fusion-based \cite{Duchon2014}. 
The first category only uses the latest sensor measurement data or weights the most recent data \cite{dey2016vision,florence2018nanomap}. In other words, these methods will not record the passed by obstacles \cite{schulman2014motion,augugliaro2012generation}. An example of this kind of method is to generate motion primitives randomly and check collision with the transformed point cloud and the trajectories \cite{lopez2017aggressive}. However, path planning directly on the latest point cloud requires the information of high quality and in full view. For UAVs with limited sensors, e.g., only one single depth camera with a narrow field of view, such methods are not applicable.

The second type is based on data fusion. Sensor data will be continuously fused into a map, usually in the form of occupying grid or distance field \cite{lau2010improved}.
Considering the map is important to plan a safe and short path in complex scenarios, many navigation frameworks are applied on a global map or local map. For building a map with the sensed environmental information, representative methods include voxel grids \cite{elfes1989using}, Octomap \cite{hornung2013octomap}, and elevation maps \cite{choi2012global}. Octomap is memory-efficient for presenting a large-scale environment and maintaining the map automatically. In \cite{lunenburg2018a}, with the point cloud raw data input, Octomap is utilized to provide the map for their proposed planning algorithms, and the experimental results are satisfactory.

\subsection{Path planning}
For path searching of UAVs, the algorithms commonly used can be classified into two categories: searching‐based or sampling-based methods. Searching‐based methods discretize the whole space into a grid map and solve path planning by graph searching. The graph can be defined in a 2D, 3D, or higher‐order state space. Typical methods include Dijkstra \cite{dijkstra1959a}, A* \cite{hart1968a}, anytime repairing A* \cite{likhachev2003ara}, JPS \cite{harabor2011online}, and hybrid A* \cite{dolgov2010path}. Dijkstra's algorithm is the root of the above methods, which searches path by utilizing an exhaustive method on all the given grids. A* improves the efficiency by setting a cost function to cut off the search away from the goal. As an improved version of the traditional A*, JPS greatly reduces its time cost without sacrificing the optimality in all the cases. However, as the path direction is constrained, the path is not the true shortest in the unconstrained 2D map.

Sampling-based methods usually do not need to discretize the space first. In the representative sampling‐based approach such as rapidly exploring random tree (RRT) \cite{lavalle2001randomized}, random and uniform sampling is performed from the space near the starting point, and the root node and child nodes are continuously connected to form a tree that grows toward the target. The RRT algorithm can effectively find a viable path, but it has no asymptotic optimality, and its search will stay at the first feasible solution. Sampling-based methods with asymptotic optimality include probabilistic road maps (PRM*) \cite{kavraki1996probabilistic}, rapid exploration of random graphs (RRG) \cite{karaman2011sampling} and RRT* \cite{karaman2011sampling}, where RRT* can make the solution converge to the global best point with the increase of samples. RRG is an extension of the RRT algorithm because it connects the new sample with all other nodes within a specific range and searches for the path after constructing the graph. Based on RRT, the method in \cite{webb2013kinodynamic} cancels the optimal control of time to ensure the asymptotic optimality of the path and kinematics feasibility. Also, a belief roadmap can be combined with RRG \cite{bry2011rapidly} to solve the problem of trajectory planning under the state uncertainty. A technique called “partial ordering” balances confidence and distance to complete the expansion graph in the confidence space.

\subsection{Trajectory and motion planning}
The trajectory planner utilizes the local obstacle information and the target point's position to plan an optimal path and a corresponding set of motion primitives within a particular time.
The artificial potential field (APF) method \cite{khatib1986real} conceives that the goal point generates an ``attractive force" to the vehicle, and the obstacle exerts a ``repulsive force". The movement of the vehicle is controlled by seeking the resultant force. Its expression is concise, but it is easy to fall into the local optimal solution. The vector field histogram (VFH) \cite{borenstein1991the} is a classical algorithm for robot navigation with a lidar, improved from the APF method. VFH will calculate the travel cost in each direction. The more obstacles in this direction, the higher the cost. The dynamic window approach (DWA) is a sampling-based method, which samples the motion primitives within the feasible space and chooses one set by ranking them with a cost function \cite{fox1997the}. The concepts in these classical methods are of the excellent reference value and enlightening significance for our method. However, for an UAV with a single depth camera flying in an unknown environment, they are not sufficient.

Inspired by the methods mentioned above, several advanced frameworks have successfully been tested in actual drone flights. For frameworks that directly obtain the motion primitives, they can also be divided into two categories: One is based on solving an optimization problem, and the other is based on sampling within the feasible state space. The former requires a wise and appropriate form to represent the UAV trajectory, e.g., Bezier curves \cite{gao2018online}, and constraints are needed to ensure the planned trajectory is collision-free. The representative works in this category are \cite{howard2008state,zhou2019robust,preiss2017trajectory}. For sampling-based methods, the evaluation function is necessary to choose the most suitable group of motion primitives as the output. A representative work is presented by \cite{mueller2015a}, in which the time cost to generate and evaluate trajectories is short enough to enable the quadrotor to even catch a falling ball.

Path planning on a local map can be conducted before the motion planning to improve the trajectory length. The local map only maintains the environment information near the UAV, while the global goal is converted to the goal in the local map. Then, the path planning algorithms introduced above can be utilized with the local map, and the motion primitives are gained based on the path. For example, Chen et al. \cite{chen2019dynamic} and Liu et al. \cite{liu2016high} adopt the minimum-jerk method to generate trajectory through the waypoints from an A* search based on the local occupancy grid map. Besides, some works studied how to allocate the flight time for a drone to fly through the waypoints. The receding horizon control policy is introduced to plan in a limited time range \cite{watterson2015safe}, and a bi-level optimization method is also effective \cite{berg2012lqg,oleynikova2018safe}. 

For the two categories, collision check is the most time-consuming part of the framework. In our previous work \cite{chen2020computationally}, we proposed a computational efficient sampling-based method with a novel collision check technique to generate collision-free trajectories on the point cloud obtained during UAV's navigation. Its loop frequency is up to 60 Hz, and it is a submodule of this project.  

In the last few year, several research works discussed how to combine the optimal global planning algorithm for static maps with the algorithm applied to real-time online replanning.
The algorithms can learn from each other's strengths and weaknesses, i.e., working out a short path and responding quickly to map changes for replanning the trajectory.
For the existence of the unknown space in the environment, several methods can be adopted: the unknown space is regarded to be freely passable in \cite{chen2016online,pivtoraiko2013incremental}, and the path is continuously adjusted as the obstacle information is updated. We call it the optimistic planner. In \cite{oleynikova2018safe,oleynikova2016continuous}, the optimistic global planner and conservative local planner are combined to ensure the safety of the aircraft. To diminish the inconsistency between the global and local planner, Tordesillas et al. \cite{tordesillas2019real} proposed a planning framework with multi-fidelity models. They run the JPS algorithm on the local slide grid map, and the constraints of motion optimization were divided into three parts according to the distance from the drone, where the most strict constraints are for the closest area.

\section{Mapping and the map planner}

\begin{figure}[thpb]
    \centering
      \includegraphics[width=0.48\textwidth]{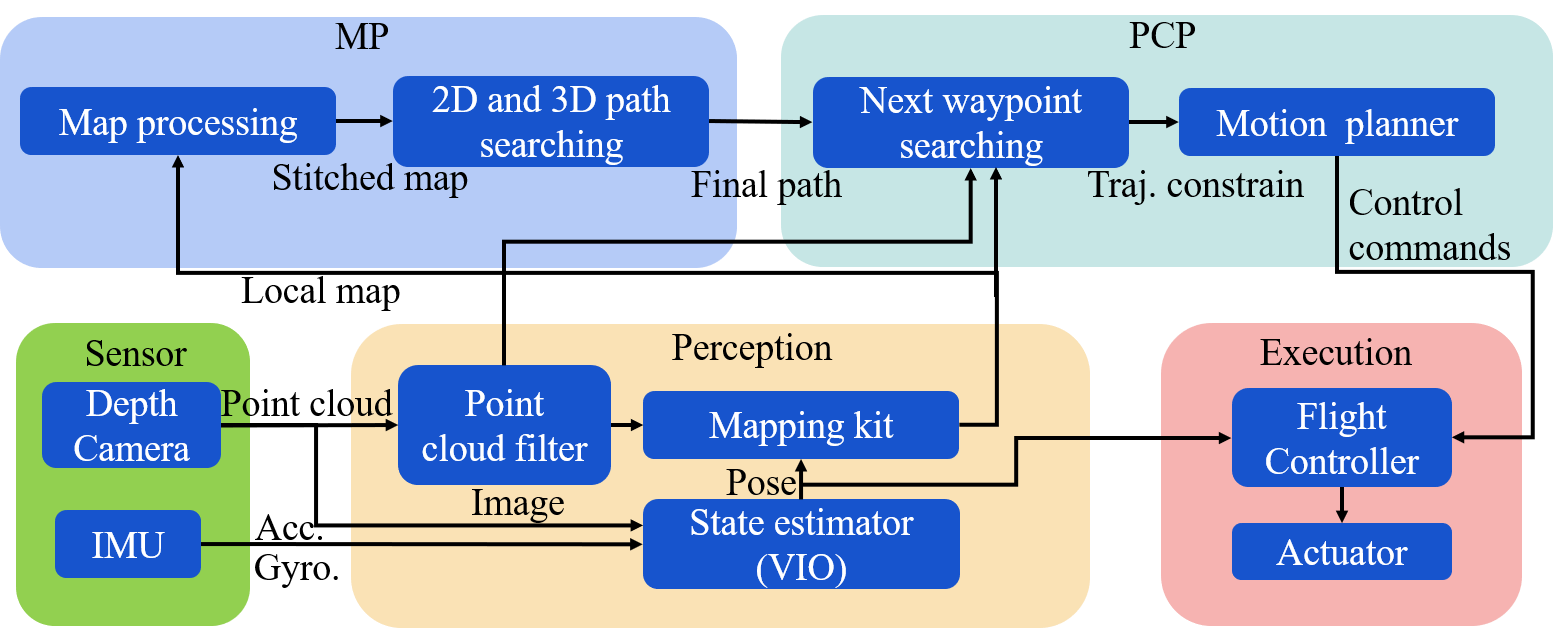}
      \caption{Architecture of our autonomous navigation system for UAVs.}
    \label{fig_archi}  
   \end{figure}

The flight system architecture is shown in Fig. \ref{fig_archi}. The MP obtains the map from the mapping kit and plans the final path as the reference, and the PCP searches the next waypoint based on the final path and optimizes the motion primitives to make the drone fly through the waypoint. 

This section primarily introduces the construction of a stitched map with two resolutions and the algorithms used for path planning in the local map (the MP). The PCP will be introduced in section \uppercase\expandafter{\romannumeral4}. Moreover, the point cloud filter for the raw sensor data preprocessing is introduced at the beginning.
\subsection{Point cloud filter}
\begin{figure}[htp]
\centering
\includegraphics[width=0.5\textwidth]{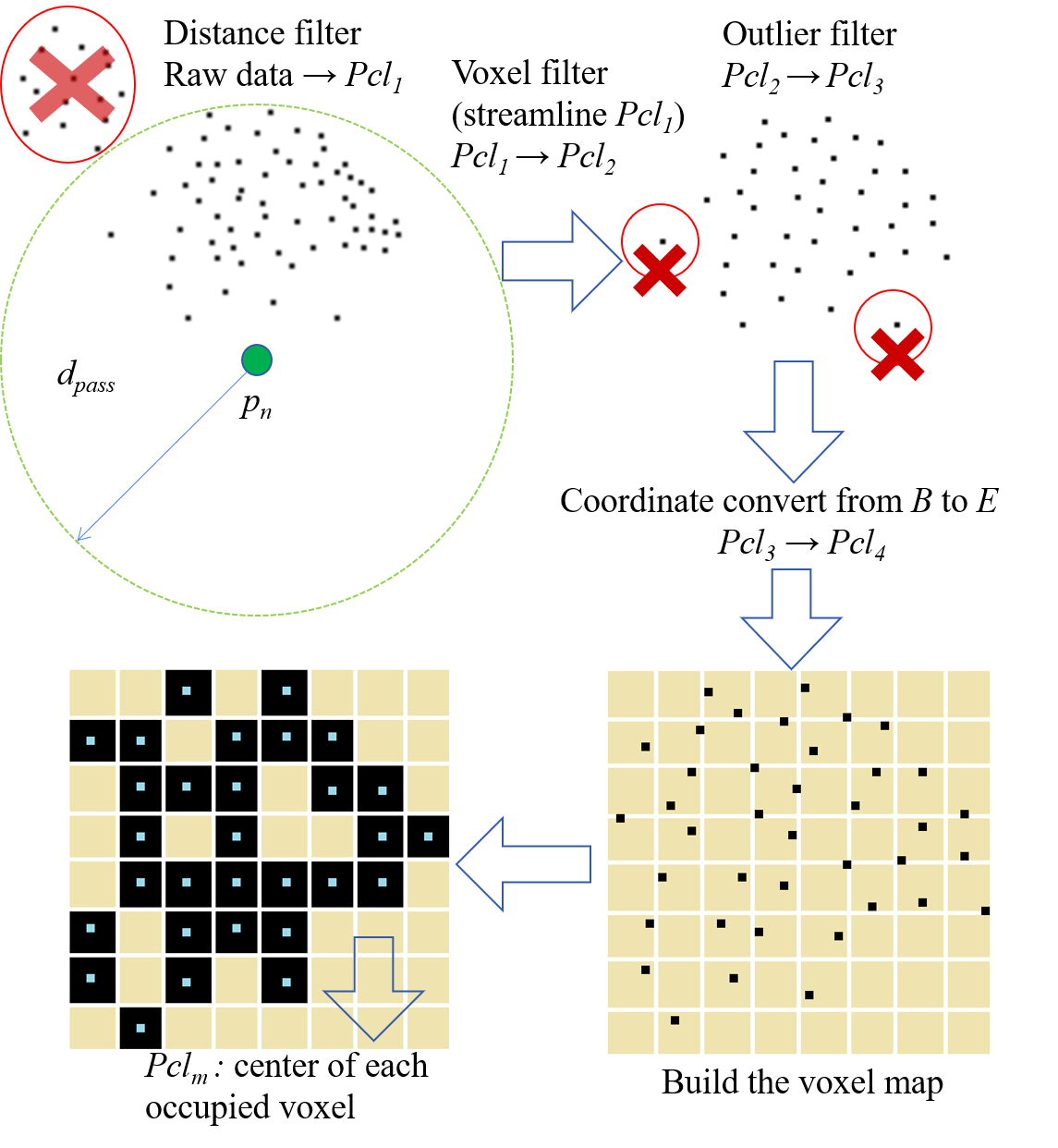}
\caption{Process for point cloud filtering, coordinate transformation, and mapping.}
\label{fig2}
\end{figure}

The dense point cloud from a real depth camera contains noise. The noise follows certain types of distributions, that is, the noise level is high when the object is far from the camera. In addition, the dense points may overburden computational procedures, such as point coordinate transformation, mapping, and planning. Moreover, the noise will mislead the mapper to mark many nonexistent obstacles on the map. Before building a map, we filter out the noise in the point cloud and keep the actual obstacle points. The process of filtering the point cloud is shown in Fig. \ref{fig2}. First, we filter the original point cloud data through the distance filter to obtain $Pcl_1$. It removes the points farther than $d_{pass}$ from the camera, which may contain too much noise. Next, a voxel filter is used to reduce $Pcl_1$ to $Pcl_2$. Furthermore, the outlier filter removes the outliers to obtain $Pcl_3$: The local point density distinguishes the outliers because the point cloud density of the noise is usually smaller. Then, we convert $Pcl_3$ into $Pcl_4$ in the Earth coordinate system and use $Pcl_4$ in the mapping kit to build and maintain a global map. Finally, the center points of occupied voxels are used as the 3D map for the collision check, referred to as $Pcl_m$. It is apparent that $Pcl_m$ well retains the basic shape of the obstacle in a more concise and tidy form. $B_E$ in (1) is the transformation matrix from body coordinate $B\!-\!xyz$ to Earth coordinate $E\!-\!X\!Y\!Z$. $c\psi$ is short for $cos(\psi)$, $s\psi$ is short for $sin(\psi)$, and $\psi$, $\phi$, and $ \theta$ are attitude angles.


At last, $Pcl_4$ is used for the map building and collision check in the PCP, and $Pcl_m$ is used for the 3D path collision check in the MP.

$$ B_{E}=\left[\begin{array}{ccc}
c \psi c \theta & s \psi c \theta & - s \theta\\
c \psi s \theta s \phi-s \psi c \phi & s \psi s \theta s \phi+c \psi c \phi & c \theta s \phi \\
c \psi s \theta c \phi+s \psi s \phi & s \psi s \theta c \phi-c \psi s \phi & c \theta c \phi
\end{array}\right] \eqno{(1)} $$

\begin{figure}[htp]
\centering
\includegraphics[width=0.5\textwidth]{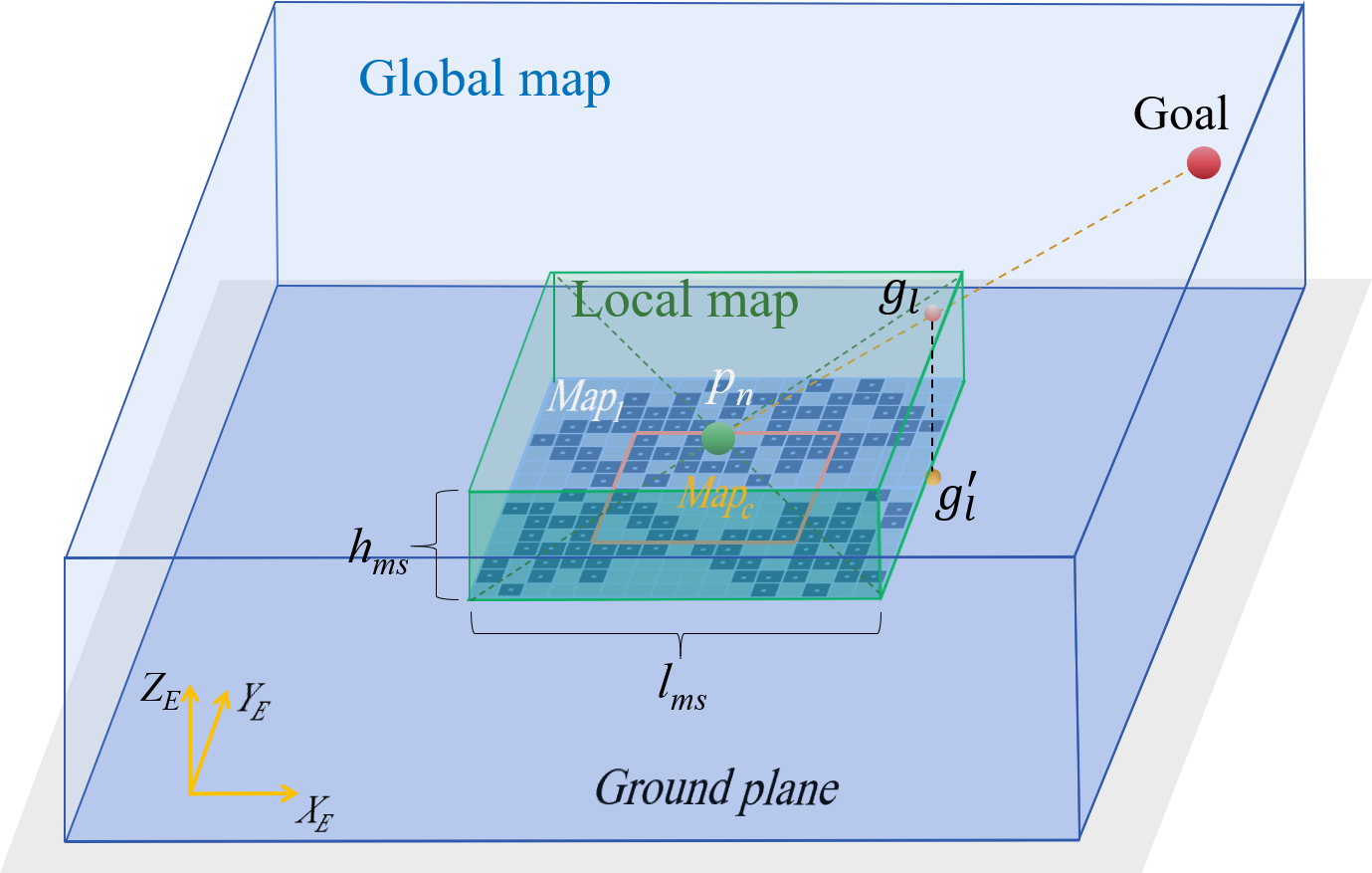}
\caption{Local and global maps. The goal projection of the local 3D map is $g_{l} \in \mathbb{R}^{3}$. $Pcl_{lm}$, while $g'_l$ is the projection of $g_l$ in the 2D local map $Map_1$, and $g'_l$ locates on the edge of $Map_1$.}
\label{fig21}
\vspace{-0.3cm}
\end{figure}

\subsection{Mapping and 2D path planning}
The mapping kit MLmapping\footnote{https://github.com/HKPolyU-UAV/MLMapping} assembled in this project is self-developed. It provides $Pcl_m$ and the projected 2D grid map for the path planning in this paper. Here, we first introduce the basic concepts of the local map. A local map is a subset of the global map and is also presented by voxels' center points. 
The space covered by the local map is a cuboid with a square bottom surface, and it has no relative rotation to the global map. As shown in Fig. \ref{fig21}, $l_{ms}$ is the square side length, and $h_{ms}$ is the local map height, which is much smaller than $l_{ms}$. The center of the local map follows the drone's current position $p_{n} \in \mathbb{R}^{3}$. $Pcl_{lm}$ represents the subset of $Pcl_{m}$ corresponding to the local map in the text below. By projecting the local map on the ground plane, the 2D grid map $Map_1$ is obtained to plan the 2D path.

To plan an optimal path on $Map_1$, JPS is one of the best choices, because it is fast and can replan the path in real time. JPS outputs the optimal path by searching a set of jump points where the path changes its direction. However, two problems arise if the path planning is performed directly on $Map_1$. First, to find a short and safe path, the local map scale should not be small. Otherwise, the optimal path on a tiny local map is more likely to end at a blind alley or differs substantially from the globally optimal path. However, a large local map is computationally expensive, and it is important to leave as much CPU resource as possible to the high-frequency PCP for safety. Second, the path planned directly on $Map_1$ is adjacent to the obstacle projection. In our framework, considering the drone frame size and flight control inaccuracies, the drone must remain at a safe distance from obstacles. Thus, the path should remain a certain distance from obstacles. The PCP will make the drone closely follow the path obtained by the MP. When the path is found to be occupied, the PCP starts to take effect. As a result, the PCP in this framework can run faster compared to that of \cite{chen2020computationally}, because the initial search direction is more likely to be collision-free.

   \begin{figure}[thpb]
      \centering
      \includegraphics[width=0.46\textwidth]{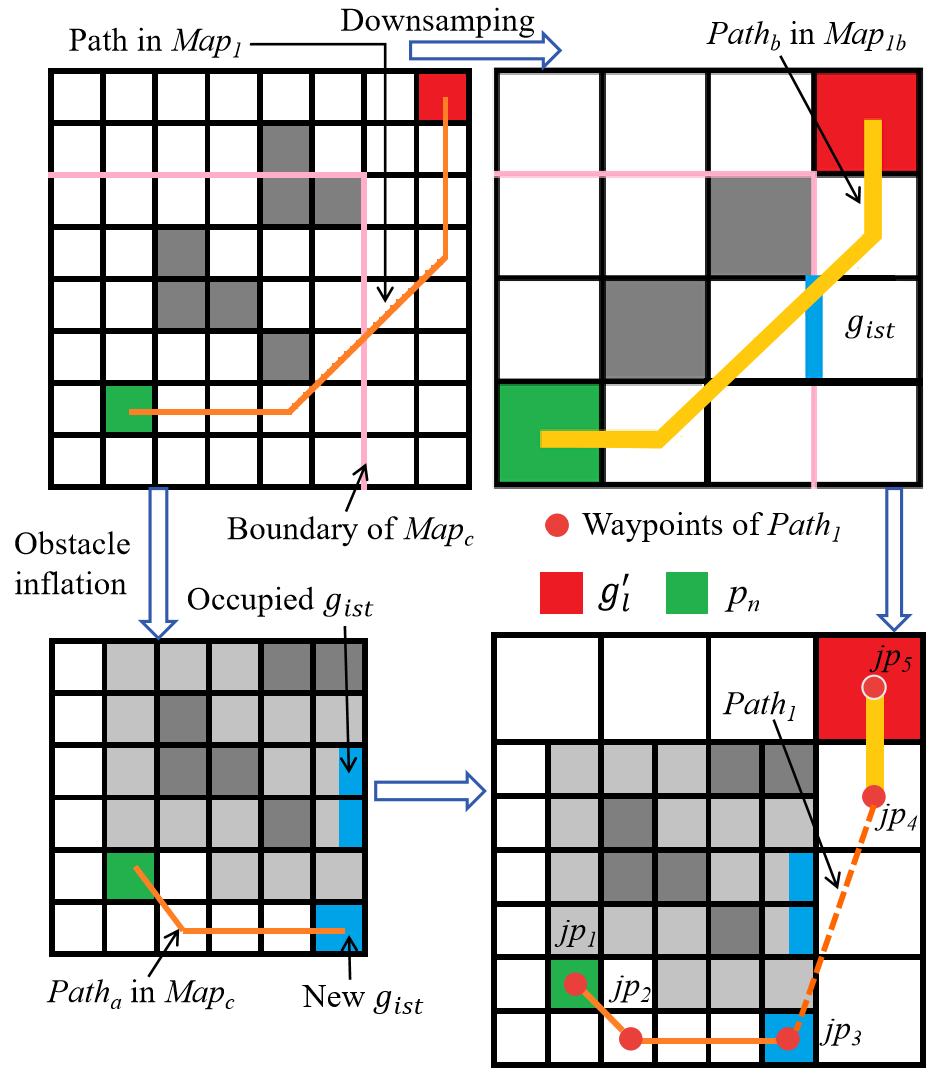}
      \caption{Partial view of the mapping process (k=3,h=2). The figures in the first row show the path in $Map_1$ and $Map_{1b}$. The figures in the second row show the path in $Map_c$, the jump points in $ Path_1$, and the stitched map. The dark gray grids indicate the obstacle, and the light gray grids are the obstacle's inflation after the convolution. The path planning start point is the center of $Map_1$ and $Map_{c}$.}
      \label{fig3}
      \vspace{-0.3cm}
   \end{figure}

In our framework, we take two measurements to address these two problems. For the first problem, we plan the path on the downsampled local map and the cast local map, respectively, and fuse the paths. We first conduct the convolution with $Map_1$ to reduce the map size and obtain a low-resolution version $Map_{1b}$ from $Map_1$. $Map_c$ is segmented from $Map_1$ afterward as the original resolution map around the drone. Then, we plan $Path_b$ on $Map_{1b}$ and find the intersection point $g_{ist}$ of $Path_b$ and the $Map_c$ boundary. The part of $Path_b$ that lies in $Map_c$ is removed. Finally, we use $g_{ist}$ as the goal point to find the path $Path_a$ in $Map_c$, and splice $Path_a$ and $Path_b$ to form a complete path $Path_1$. The grid size of $Map_{1b}$ positively correlates with the map size, so the time cost of 2D path planning can be controlled.


For the second problem, after we have obtained $Map_c$, we first perform an expansion operation on the obstacles in $Map_c$. Using a convolution kernel to convolve the binary matrix corresponding to $Map_c$, the blank area next to the obstacle in the map can be marked as an obstacle so that each point on $Path_1$ maintains a certain distance from true obstacles.

$$Map'_{1} = \left[\begin{array}{cc} [Map_{1}]_{i \times j}&[\textbf{0}]_{(i+s) \times s}\\
{}[\textbf{0}]_{s \times j}&
\end{array}\right]\eqno{(2)}$$
$$Map'_{c} = \left[\begin{array}{ccc} &[\textbf{0}]_{k \times (n+2k)}&\\
{}[\textbf{0}]_{(m+2k) \times k} & [Map_{c}]_{m \times n} & [\textbf{0}]_{(m+2k) \times k}\\
 &[\textbf{0}]_{k \times (n+2k)}& \end{array}\right]\eqno{(3)}$$
$$Map_{c} = Sgn(Conv_{1}([Map'_{c}]_{(m+2k)\times (n+2k)},I_{k\times k}))\eqno{(4)}$$
$$Map_{1b}=\langle Conv_{h}([Map'_{1}]_{(i+s)\times(j+s)},\dfrac{I_{h\times h}}{h^{2}})\rangle \eqno{(5)}$$
$$h=\langle\dfrac{ij}{2mn}\rangle \ (ij>3mn \ \&\ (i+s)\ \text{is divisible by}\ h)\eqno{(6)}$$

Equations (2)-(5) show the calculation of the downsampling and inflation. $i$ and $j$ denote the size of the original $Map_1$ ($i=j$), and $m$ and $n$ denote the size of the cut map $Map_c$ ($m=n$). We use $[\ ]$ to present a matrix, and the subscript of the matrix denotes its size. $[\textbf{0}]$ indicates the zero matrix. $s$ is the line and column number for zero padding for $Map_1$. $h$ and $k$ are the convolution kernels' size for map downsampling and obstacle inflation, respectively. $Sgn()$ is a function that returns the sign matrix corresponding to each element in the input matrix. The sign matrix is used as the binary map with two types of elements: 0 and 1. $Conv()$ indicates the convolution, and it inflates the occupancy grids on the map or downsamples $Map_1$. Its subscription indicates the step size for the kernel sliding, and the second element is the convolution kernel. $\langle \rangle$ is for rounding the number to the nearest integer. If a matrix is in $\langle \rangle$, it rounds each element in the matrix. (4) represents the obstacle inflation process, and (5) is for the map downsampling. Fig. \ref{fig3} illustrates the map processing and path planning intuitively. The deep gray grids represent the obstacles, and the light gray grids are the inflation of obstacles after the convolution. $g_{ist}$ is represented in blue on the maps. When $g_{ist}$ is occupied after the inflation, we find the nearest free grid on the map edge as the new $g_{ist}$. The calculation of $h$ is introduced in (6), and $i,\ j,\ m,\ n,\ s$ should meet the conditions in the bracket.

\subsection{Improved 2D path}

In section \uppercase\expandafter{\romannumeral3}.B, a path $Path_{1}(jp_{1},jp_{2},...)$ on a plane parallel to the ground plane $XY$ is found using the JPS method in the hybrid map of $Map_{1b}$ and $Map_{c}$. However, in some cases, it is not the shortest path in the plane, as search directions of waypoints can only be a multiple of $45^{\circ}$. We can further optimize the original path by deleting the redundant waypoints. For example, in Fig. \ref{fig32}, the red path is the original path, the green path is the improved path, and $jp_2$ and $jp_4$ are deleted. The deleting process can be written in Algorithm \ref{alg21}. $ti$ is the iteration number, $jp_{ck}$ is the $ck\text{th}$ point in $Path_1$, and the same for $jp_{ti}$. We connect the third point in the original JPS path with the first point and check if the line collides with the occupied grid in the map. If it does not collide, the point before the checked point in the waypoint sequence of the original JPS path is deleted. The first and third points can be directly connected as the path. Then, the next point will be checked until all the point pairs (the two points are not adjacent) from $Path_1$ are checked, and all excess waypoints in $Path_1$ are removed.

\begin{algorithm}[htp]
\caption{Optimize the original JPS path} 
\label{alg21}
\begin{algorithmic}[1]
\FOR{$jp_{ck}$ in $Path_1$ ($ck$ is the iteration number):}
\STATE $ti=ck+2$
\WHILE{$ti<len(Path_{1})$ and $len(Path_{1})>2$}
\IF{$\overline{jp_{ck}jp_{ti}}$ does not collide with the occupied grids in the 2D map:}
\STATE $ti = ti-1$, delete $jp_{ti}$ from $Path_1$
\ENDIF
\STATE $ti=ti+1$
\ENDWHILE
\ENDFOR
\end{algorithmic}
\end{algorithm}  	

\subsection{Shorter 3D path searching}
\begin{figure}[htb]
      \centering
      \includegraphics[width=0.50\textwidth]{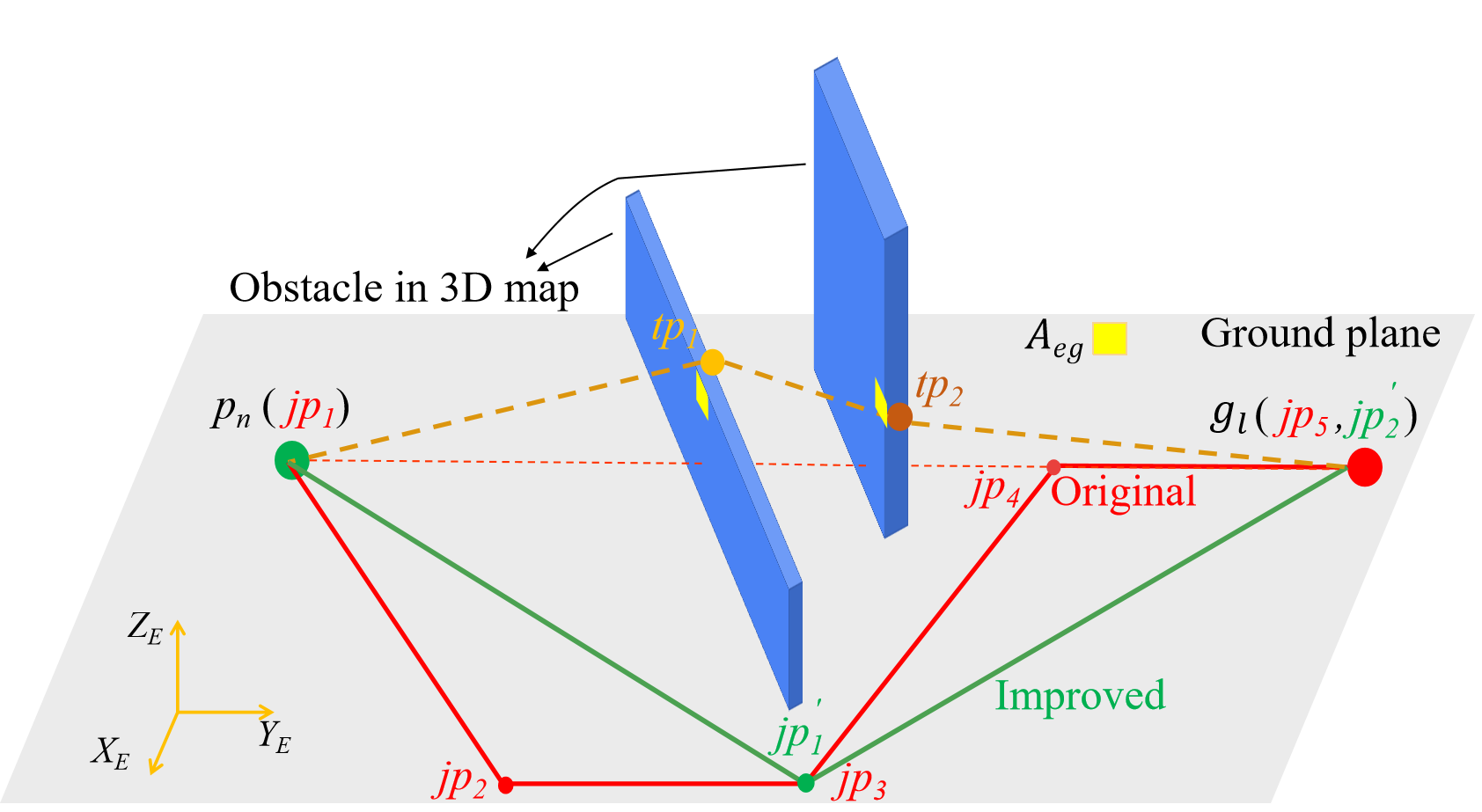}
      \caption{A scenario where the 3D path is much shorter than the improved 2D path.}
      \label{fig32}
   \end{figure}

\begin{figure}[htb]
\centering
\subfigure[]{
      \includegraphics[width=0.215\textwidth]{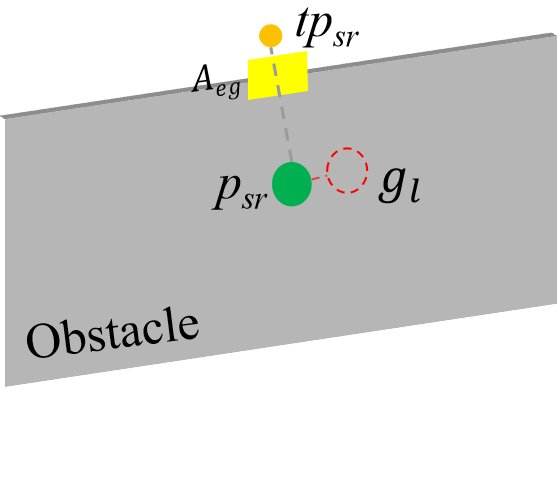}}
      \centering
\subfigure[]{\includegraphics[width=0.26\textwidth]{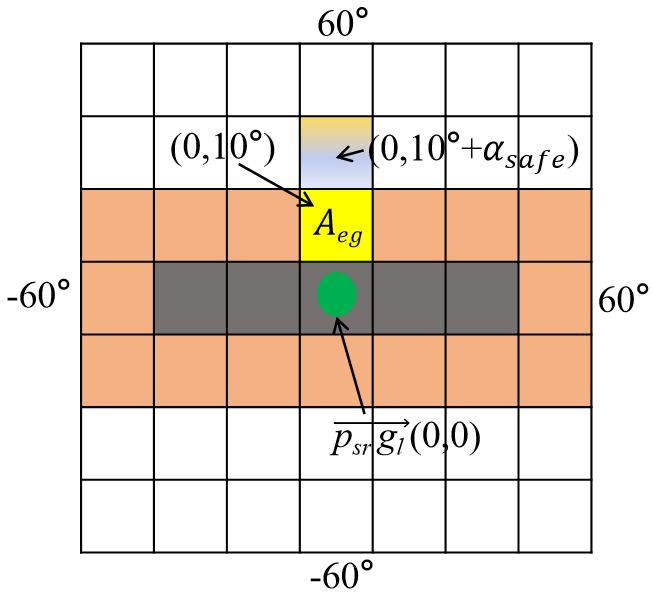}}
      \caption{(a): A wall stands between the drone $p_n$ and the local goal $g_l$, (b): The discrete angular graph for (a), $\alpha_{res}=20^\circ$. The orange grids are the edge grid of the obstacle.}
      \label{fig31}
   \end{figure}

After an improved 2D path is found, we notice an obviously shorter 3D path in some scenarios. For example, to avoid a wall, which has large width but limited height, flying above the wall is better than flying over a bypass from right or left. To search for a shorter 3D path with light computation, a generalized method DAGS for all environments is described in Algorithm \ref{alg31}, Fig. \ref{fig32}, and Fig. \ref{fig31} in detail. It is composed of two rounds of search, and each round determines one straight line segment to compose the 3D path. $Pcl_{lm}$ is divided into two parts: One part is denoted as $Pcl_{lm\!-\!1}$ and is composed of the points whose distance to $p_n$ is smaller than $\overline{p_{n}jp'_{1}}$. Another part $Pcl_{lm\!-\!2}$ consists of the remaining points. As shown in Fig. \ref{fig32}, the first segment is $\overline{p_{n}tp_{1}}$, the second segment is $\overline{tp_{1}tp_{2}}$, and $p_{n}\!-\!tp_{1}\!-\!tp_{2}\!-\!g_{l}$ represents the shorter 3D path. $\alpha_{res}$ is the angular resolution of the discrete angular graph. $A_g(\alpha_{g1},\alpha_{g2})$ is the angular part of the spherical coordinates of $g_l$, and the origin of the spherical coordinate system is $p_{sr}$ for each search round. $min()$ is a function that returns the minimal value of an array.

Here, the procedure for the first round of the search is briefly introduced, and the second round is basically identical. First, the discrete angular graph is built by Algorithm \ref{alg31}, line 3-8, as shown in Fig. \ref{fig31}(b). $\lfloor \ \rfloor$ returns the integer part of each element of the input. The angular coordinate in the graph is the direction angle difference between the goal $g_l$ and any point in the space. The colored grids represent all the discrete angular coordinates $A_{mid-d}$ corresponding to the input point cloud. Then, the relative direction angle $A_{eg}$ for $\overline{p_{n}tp_{1}}$ is found (line 9), which has the minimal angle difference with $\overrightarrow{p_{n}g_l}$ (the yellow grid in Fig. \ref{fig32} and Fig. \ref{fig31}). Next, the length of path segment $\overline{p_{n}tp_{1}}$ is determined in line 10, and the direction angle of this path segment in $E\!\!-\!\!X\!Y\!Z$ is calculated by line 11. $\alpha_{safe}$ is the angle increment to make the path segment remain a safe distance from obstacles. Finally, the coordinate of the endpoint of this path segment is calculated in line 13. If the 3D path is found by Algorithm \ref{alg31}, it is compared to the optimized $Path_{1}$, and the path with the shortest length is denoted as $Path_{fnl}$. Subsequently, the drone follows $Path_{fnl}$, and the MP is suspended until $Path_{fnl}$ collides with the obstacles in the updated map.

\begin{algorithm}[htp]
\caption{DAGS method} 
\label{alg31}
\begin{algorithmic}[1]
\FOR{$sr$ in \{1,2\} ($sr$ is the searching rounds number)}
\STATE If $sr=1$, $p_{sr} = p_{n}$, otherwise $p_{sr} = tp_{1}$
\STATE {The angular coordinate of $\overrightarrow{p_{sr}g_{l}}$ $\rightarrow$ $A_{g}(\alpha_{g1},\alpha_{g2})$}
\FOR{each point $p_{mi}$ in $Pcl_{lm-sr}$:}
\STATE The angular coordinate of $p_{mi}$ $\rightarrow$ $A_{mi}$
\STATE $A_{mig}=A_{mi}-A_{g}$
\STATE Discretize $A_{mig}$, $A_{mig-d}=\lfloor A_{mig}/\alpha_{res}\rfloor$, build the discrete angular graph with $A_{mig\!-\!d}$
\ENDFOR
\STATE The edge of all $A_{mig-d}$ in the angular graph $\rightarrow$ $A_{eg-all}$, $A_{eg} \subset  A_{eg-all}$ and $\|A_{eg}\|_{2}=min(\|A_{eg-all}(1)\|_{2},\|A_{eg-all}(2)\|_{2},...)$
\STATE Points in $Pcl_{lm-sr}$ corresponding to $A_{eg}$ $\rightarrow$ $Pcl_{eg}$, $p_{eg} \subset Pcl_{eg}$ and $p_{eg}$ has the maximal distance to $\overline{p_{sr}g_{l}}$, $l_{tp}=\overline{p_{n}p_{eg}}$
\STATE Get the direction angle of $\overrightarrow {p_{sr}tp_{sr}}$, $A_{tp} =(\|A_{eg}\|_{2}+ \alpha_{safe})\dfrac{A_{eg}}{\|A_{eg}\|_{2}}$, $\alpha_{safe}=arcsin(r_{safe}/l_{tp})$

\STATE $tp_{sr} = p_{sr} + l_{tp}(cos(A_{tp}(1)),sin(A_{tp}(1)), sin(A_{tp}(2)))$
\ENDFOR
\end{algorithmic}
\end{algorithm}  	

\section{Trajectory planning on the point cloud}

This section will introduce how the goal $g_n$ is generated from $Path_{fnl}$ for the PCP's current step $n$ and how the PCP outputs the final motion primitives. First, the discrete angular search (DAS) method specifies the safe waypoint $w_{pn} \in \mathbb{R}^{3}$ in free space, which the drone should traverse. The motion planner solves the optimization equation to make the drone pass through $w_{pn}$ under the given motion constraints. The PCP also includes an additional safety measure to ensure that no collision will occur, which works when no $w_{pn}$ can be found in an emergency.

\subsection{Review of DAS}
The PCP in this paper is an improved version of our previously proposed trajectory planner based on the DAS method \cite{chen2020computationally}. Thus, briefly introducing it helps to understand the contributions in this section.

\begin{figure}[thpb]
\centering
\includegraphics[width=0.47\textwidth]{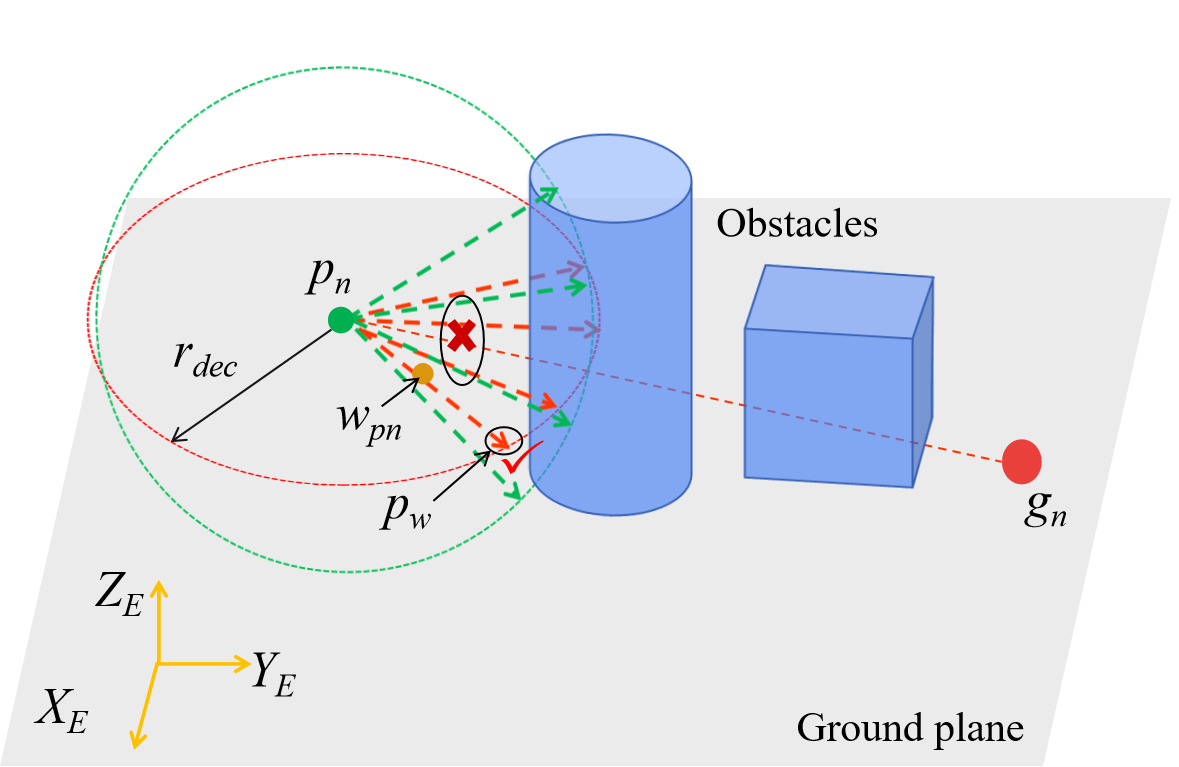}
\hfill
\caption{Illustration of the DAS process. The red dash circle and red arrows are for the search on the ground-parallel plane, while the green ones are for the vertical plane. In this figure, the collision check for the first line of the second search round is successful (denoted as $\overline{p_{n}p_{w}}$).}
\label{fig61}
\vspace{-0.6cm}
\end{figure}

In \cite{chen2020computationally}, we first search a waypoint $w_{pn}$ near the drone as the constraint for the motion planning, and a quadratic polynomial curve is optimized as the final trajectory in the motion planning. Motion planning is for solving a nonlinear optimization problem. The trajectory traverses $w_{pn}$ within a small, predetermined distance error, and the corresponding motion primitives are within the kinematic constraints. Moreover, the real trajectory between $p_n$ and $w_{pn}$ can be proved collision-free. As shown in Fig. \ref{fig61}, $g_{n}$ is the goal point for the current step, a cluster of line segments fanned out in the direction of $\overrightarrow{p_{n}g_{n}}$, and these line segments have a common starting point $p_n$ and the same length $r_{det}$. In this paper, $g_{n}$ is determined by the planned path $Path_{fnl}$. $r_{det}$ is the point cloud distance threshold, and the collision check only considers the points within distance $r_{det}$ from $p_{n}$. The two symmetrical lines about $\overrightarrow{p_{n}g_{n}}$ on the plane parallel to the ground plane are first checked to determine if they collide with obstacles (minimal distance smaller than $r_{safe}$). $r_{safe}$ is an important parameter for safety, which is a function of the preassigned maximal speed $v_{max}$ and acceleration $a_{max}$. Then, the two symmetrical lines about $\overrightarrow{p_{n}g_{n}}$ in the vertical plane are checked. Fig. \ref{fig61} shows that when the first round of the search fails, another round with a greater angle difference is conducted until a collision-free line $\overline{p_{n}p_{w}}$ is found. $w_{pn}$ is on $\overline{p_{n}p_{w}}$, and $\overline{p_{n}w_{pn}}$ should satisfy the safety analysis in \cite{chen2020computationally}).

When the drone encounters a suddenly intruded obstacle in the sensor detection range, the PCP plans a safe trajectory in 0.02 s before the map and $Path_{fnl}$ are updated.

\subsection{Connection between the PCP and MP}

For the PCP, an updated goal point $g_{n}$ is always required at every step $n$. The direction $\overrightarrow{p_{n}g_{n}}$ is the initial search direction. If this direction does not collide with any obstacles, the planned trajectory will head to $g_{n}$ directly.

The final path $Path_{fnl}=[pt_1,pt_2,...,g_l]$ is received from the MP ($Path_{fnl}$ dose not include $p_n$), and an optimization problem (8) is designed to find the current goal $g_{n}$. It is designed to make the final trajectory smooth by sliding $g_{n}$ continuously. If $pt_1$ is simply assigned as $g_{n}$, $g_{n}$ will jump to $pt_{2}$ as the drone approaching $pt_1$. This result may cause $w_{pn}$ to also jump with $g_{n}$ and cannot be reached within the drone's kinematic constraints. The drone should start to turn earlier to avoid a violent maneuver, which leads to a greater control error and undermines safety. The endpoint of the planned trajectory remains near $Path_{fnl}$ under the premise of safety assurance.

\vspace{-0.3cm}
$$\begin{aligned}
\min _{v_{t}} \quad & \|v_{t}\dfrac{\|v_{0}\|_2}{\|v_{t}\|_2}-v_{0}\|_{2}+\|v_{t}-\kappa_{1}a_{1}\|_{2} + \|v_{t}-\kappa_{2}a_{2}\|_{2} \\
\text {s.t.} \quad & a_1=pt_{1}-p_{n},\ a_2=pt_{2}-p_{n}, \ 
v_{t}= \overrightarrow{p_{n}g_{n}}
\end{aligned} \eqno{(8)}$$

\begin{figure}[thpb]
    \centering
      \includegraphics[width=0.38\textwidth]{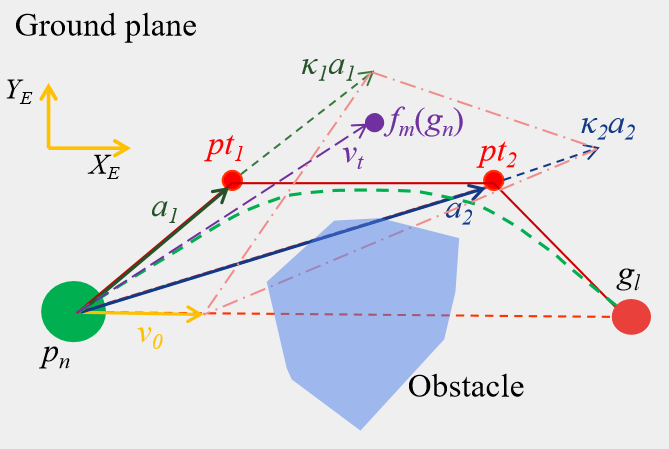}
      \caption{Geometric illustration of the analytical solution of (8). The pink, dashed line marks the Fermat triangle.}
    \label{fig4}  
   \end{figure}

In (8), the three components are the acceleration cost, the cost of $pt_1$, and the cost of $pt_2$. $p_n$ is the current position of the drone, and $v_0$ is the current velocity. $v_t$ presents the initial search direction of the local planner. $r_{det}$ is the search range radius for the DAS. $\kappa_1$ and $\kappa_2$ are the weight factors for adjusting the influence of $pt_1$ and $pt_2$ on $v_t$, and $\kappa_1$ is much larger than $\kappa_2$. Fig. \ref{fig4} intuitively demonstrates the initial search direction $v_t$ for the PCP, drone position, and waypoints on $Path_{fnl}$. The green, dashed line displays the rough shape of the trajectory if the PCP does not check for a collision and $w_{pn}$ is always on $\overline{p_{n}g_{n}}$. We can see from (8) and Fig. \ref{fig4} that as the drone approaches the next waypoint $pt_1$, the influence on $g_{n}$ from $a_2$ overwhelms $a_1$. When the drone is far from $pt_1$ and $pt_2$, $a_1$ is the governing influence factor of $v_t$. We can also reduce the difference between the trajectory and $Path_{fnl}$ by regulating $\kappa_1$ and $\kappa_2$. If only one waypoint $g_l$ remains in $Path_{fnl}$, $pt_1=pt_2=g_l$.

Solving a nonlinear optimization problem, such as (8), is computationally expensive. From a geometric point of view, the nature of (8) is to find a point ($v_t$) in the space with the minimal total distance between three fixed points ($\kappa_{1}a_{1}+p_{n}$,\ $\kappa_{2}a_{2}+p_{n}$,\ $v_{0}+p_{n}$). The triangle composed of these three points is called a Fermat triangle. It can be solved by locating the \textbf{Fermat point} $f_m$ of the triangle, as shown in Fig. \ref{fig4}, and $g_{n}=f_{m}$. The calculation of $f_m$ is illustrated in (9). First, the plane coordinates $(x_{f1},y_{f1})$, $(x_{f2},y_{f2})$, and $(x_{f3},y_{f3})$ in the plane $P_{fm}$ for the three points are determined ($P_{fm}$ is the plane defined by the Fermat triangle). $S_{fm}$ is the area of the triangle. $l_{1}$ is the length of the side opposite to the triangle point $(x_{f1},y_{f1})$, and so on. $f'_{m}(x_{fm},y_{fm})$ is the coordinate of $f_m$ in plane $P_{fm}$. Finally, $f'_{m}$ is converted to $E\!-\!X\!Y\!Z$ to obtain $f_m$.


$$ \begin{aligned}
& x_{fm}= (\sum_{i=1}^{3}x_{fi}(4S_{fm}+\sqrt{3}l^{2}_{i})+g(y))/f_{Sl}   \\
   & y_{fm}=(\sum_{i=1}^{3}y_{fi}(4S_{fm}+\sqrt{3}l^{2}_{i})+g(x))/f_{Sl}\\
   & g(x)= [x_{f1},x_{f2},x_{f3}][l^{2}_{3}-l^{2}_{2},l^{2}_{1}-l^{2}_{3},l^{2}_{2}-l^{2}_{1}]^{T}\\
   &  g(y)= [y_{f1},y_{f2},y_{f3}][l^{2}_{3}-l^{2}_{2},l^{2}_{1}-l^{2}_{3},l^{2}_{2}-l^{2}_{1}]^{T}\\
   & f_{Sl}= 12S_{fm}+\sqrt{3}(l^{2}_{1}+l^{2}_{2}+l^{2}_{3})
   \end{aligned}\eqno{(9)}$$



          

\subsection{Improvements on the PCP}
\subsubsection{Streamline and sequence the input point cloud}
For the input point cloud $Pcl_{r}$ of the PCP, $Pcl_{r}=Pcl_{mr} \cup Pcl_{4r}$. $Pcl_{4r}$ is the subset of $Pcl_4$, $Pcl_{mr}$ is the subset of $Pcl_m$, and their distance to $p_n$ is within $r_{det}$. Not only $Pcl_{4r}$ but also $Pcl_{mr}$ is used as the input point cloud because the camera field of view (FOV) is narrow. If only $Pcl_{4r}$ is checked for the collision, the drone may still collide with the obstacles outside the FOV.

In \cite{chen2020computationally}, we find that the execution time of the trajectory planner is highly relevant to the size of the input point cloud. The collision check accounts for a large proportion of the total time cost. Every point in $Pcl_{r}$ is checked for a collision in the loops. Once a point $p_{1st}$ in $Pcl_{r}$ is first found to collide with the detecting line segment, the collision check loops are terminated. Therefore, we hope to find $p_{1st}$ in a more computation-efficient way to reduce the time cost.

By analyzing a large amount of recorded data of the PCP in simulation and hardware tests, we found the following statistical laws. $p_{1st}$ has a larger probability to appear in the part that is closer to $p_n$ and $\overline{p_{n}g_{n}}$ ($Pcl_{r}$ is first sorted in order of increasing distance to $p_n$). $d_{ft}$ is the farthest distance from $p_n$ to the points in $Pcl_{r}$. For approximately $89.6\%$ of all the recorded $p_{1st}$, $\overline{p_{n}p_{1st}} \leq 0.5d_{ft}$. For approximately $81.3\%$ of $p_{1st}$, the angle $\angle p_{1st}p_{n}g_{n} \leq 90^{\circ}$. Thus, the part of $Pcl_{r}$ that is out of the highlight range ($\overline{p_{n}p_{1st}} \leq 0.5d_{ft}$, and $\angle p_{1st}p_{n}g_{n} \leq 90^{\circ}$) can be streamlined. In Algorithm \ref{alg41}, we streamline $Pcl_{r}$ to remain within at most $n_{use}$ points of it. $len()$ returns the list size. $D_{pn}$ is the list that stores the distance between the points in $Pcl_{4r}$ and $p_n$. The point that is more likely to collide is stored in list $Pcl_{use}$ to have a higher priority for being checked (line 3-4). When the number of points in $Pcl_{use}$ is more than $n_{use}$, $Pcl_{use}$ is evenly spaced to limit its size. The limited size of $Pcl_{use}$ reduces and stabilizes the time cost for the collision check. In addition, if $n_{use}$ and $r_{safe}$ are reasonable, safety is not compromised in extensive simulation and hardware tests.


\begin{algorithm}[htp]
\caption{Streamline the sorted $Pcl_{4r}$} 
\label{alg41}
\begin{algorithmic}[1]
\FOR{$pc_j$ in $Pcl_{4r}$ ($j$ is the iteration number):}
\IF {($D_{pn}(j)\le 0.5d_{ft}$ or $\angle p_{1st}p_{n}g_{n} \leq 90^{\circ}$):}
\STATE Put $pc_j$ in list $Pcl_{use}$
\ENDIF
\ENDFOR
\IF{$len(Pcl_{use}) > n_{use}$:} 
\STATE Remove $len(Pcl_{use})-n_{use}$ points in $Pcl_{use}$ randomly
\ELSE
\STATE Choose $n_{use}-len(Pcl_{use})$ points from $\complement_{Pcl_{4r}}Pcl_{use}$ randomly, add them into $Pcl_{use}$
\ENDIF
\end{algorithmic}
\end{algorithm}  	
\vspace{-0.3cm}
\subsubsection{Improve the motion primitives generation efficiency}

After obtaining the next waypoint $w_{pn}$, the next step is to calculate the motion primitives and send the command to the flight controller. Sending $w_{pn}$ directly as the control command may cause the flight to be unstable, and the acceleration magnitude may exceed $a_{max}$. Because $w_{pn}$ may vary significantly between two continuous motion planning steps, the point cloud quality is harmed when the drone acceleration magnitude is too large. In addition, the position commands cannot control the speed. To ensure that the aircraft can fly within its kinematic limits and reach the next waypoint, the motion primitives are generally obtained by solving an optimization problem. Previously \cite{chen2020computationally}, we considered the flight time to reach $w_{pn}$ the optimization variable. However, we found the solver may fail in the given number of iteration steps in some cases. The solving success rate with an error tolerance $10^{-3}$ is approximately $83.7\%$ within 40 steps. When the solver fails in several continuous PCP steps, the planned trajectory deviates considerably from $w_{pn}$, and the drone is very dangerous. Furthermore, considering the requirement of low time cost for real-time computing, the maximal number of iteration steps should be limited. The time variable increases the problem complexity, and the flight controller does not require it. Therefore, the flight time can be removed from the optimization variables and the optimization strategy (7) is proposed. It is slightly different from that of \cite{chen2020computationally}, to improve the success rate and time cost.

\vspace{-0.5cm}
$$\begin{aligned}
\min _{a_{n}} & \left\|a_{n}\right\|_{2}^{2}+\eta_{1} \| \overrightarrow{w_{pn}p_{n+1}}\|_{2}+\eta_{2} \dfrac{\|\overrightarrow{p_{n}p^{*}_{n+1}}\times \overrightarrow{p^{*}_{n+1}w_{pn}}\|_{2}}{\|\overrightarrow{p_{n}w_{pn}}\|_2}\\
\text {s.t.} \ \  & v_{n} =\dot{p}_{n},\ a_{n} =\dot{v}_{n} \\
&\left\|v_{n+1}\right\|_{2} \leq v_{\max },\ \left\|a_{n}\right\|_{2} \leq a_{\max }\\
&v_{n+1}=v_{n}+a_{n} t_{avs}\\
&p_{n+1}=p_{n}+v_{n} t_{avs}+\frac{1}{2} a_{n} t_{avs}^{2}\\
&p^{*}_{n+1}=p_{n}+2v_{n} t_{avs}+2 a_{n} t_{avs}^{2}
\end{aligned} \eqno{(7)}$$
\vspace{-0.0cm}

In the revised optimization formula, we fix the trajectory predicting time to $t_{avs}$. $t_{avs}$ is the average time cost of the last 10 executions of the PCP. The endpoint constraint is moved to the objective function. The endpoint of the predicted trajectory need not coincidence with $w_{pn}$. Because the execution time of the PCP is always much smaller than the planned time to reach $w_{pn}$, before the drone reaches $w_{pn}$, a new trajectory is generated, and then the remainder of the formerly predicted trajectory is abandoned. Predicting only the trajectory between the current step to the next step of the PCP is sufficient. Therefore, minimizing the distance between the trajectory endpoint at $t_{avs}$ and $w_{pn}$ is reasonable. Given that the current step run time of the PCP may exceed $t_{avs}$, the distance from the trajectory endpoint at $2t_{avs}$ to $\overline{p_{n}w_{pn}}$ should also be optimized. The subscript $n$ presents the current step in a rolling process of the PCP. $v_{n} \in \mathbb{R}^{3}$ and $a_{n} \in \mathbb{R}^{3}$ are the current velocity and acceleration of the drone. $v_{max}$ and $a_{max}$ are the kinematic constraints for speed and acceleration, respectively, and $v_{n+1}$, $p_{n+1}$, and $p_{n+1}$ are calculated using the kinematic formula. $\eta_1$, $\eta_2$ are the weight factors for the trajectory endpoint constraint.

After the modification, the success rate with error tolerance $10^{-3}$ within 20 steps is increased to $99.8\%$, and no dangerous trajectory deviation from $w_{pn}$ can be detected. The time cost of the motion planning and safety of the PCP is greatly improved.   

\subsubsection{Safety backup plan}

On some occasions, such as when the obstacles are too dense or an obstacle suddenly appears near the drone (distance is smaller than $r_{safe}$), DAS may fail to find a feasible direction. To solve this problem, the minimum braking distance $d_{bkd}=\dfrac{\|v_{n}\|^{2}_{2}}{2a_{max}}$ at current velocity $v_{n}$ is introduced. It is smaller than $r_{safe}$ by setting the appropriate maximum acceleration constraint $a_{max}$ and velocity constrain $v_{max}$ ($\|v_{n}\|_{2} \leq v_{max}$). If the minimum distance from the drone to obstacles is greater than $d_{bkd}$, the search direction having the maximum distance to the obstacles is chosen (although the distance is smaller than $r_{safe}$). Otherwise, the drone brakes immediately and flies back to the position at the former PCP step, and the chosen search direction of the former step will not be considered after the drone has flown back in place. This measure is called the ``safety backup plan.''

\subsection{The whole framework}
\begin{algorithm}[htp]
\caption{our proposed framework} 
\label{alg1}
\begin{algorithmic}[1]
\WHILE{true: (Thread 1)}
\STATE Filter the raw point cloud data, output $Pcl_4$
\ENDWHILE
\WHILE{true: (Thread 2)}
\STATE Build a global 3D voxel map ${Pcl_{m}}$, and project it on the ground to obtain $Map_1$
\ENDWHILE
\WHILE{the goal is not reached: (Thread 3)}
\IF{the shorter 3D path has not been found or it collides with the updated $Pcl_m$:}
\STATE Apply convolution on ${Map_1}$, find the 2D path $Path_1$ on the stitched map.
\STATE Try to find a shorter 3D path, output $Path_{fnl}$
\ENDIF
\ENDWHILE
\WHILE{the goal is not reached: (Thread 4)}
\STATE Calculate the goal $g_{n}$ from $Path_{fnl}$
\STATE Find the waypoint $w_{pn}$ by DAS method
\IF{$\textbf{found a feasible waypoint:}$} 
\STATE Run the motion planner to get motion primitives
\ELSE 
\STATE Run the safety backup plan, and go to line 14
\ENDIF
\STATE Send the motion primitives to the UAV flight controller
\ENDWHILE
\end{algorithmic}
\end{algorithm}  	
To summarize, Algorithm \ref{alg1} shows the overall proposed framework. The two planners (MP and PCP) are designed to run in ROS parallelly and asynchronously because of their large difference in operation time and share all the data involved in the calculation via the ROS master node. In addition, the point cloud filter and the mapping kit are run in parallel on different threads.

\section{Test Results}

In this section, the static tests and real-time flight tests are introduced to validate the effectiveness of the methods in our proposed framework.

\subsection{Algorithm performance static test}
Our detailed algorithms and methods are designed to obtain an undiminished or much better path planning performance with a decreased or slightly increased time cost. To prove our proposed algorithms' effectiveness, we first individually test them offline with static data input. This approach can avoid the influence from the fluctuations in computing performance caused by other simultaneously running algorithms when one algorithm is analyzed. Moreover, the data can be customized, so the tests are more effective and targeted. In this subsection, all the time costs are measured on a personal computer with an Intel Core i7-8565U 1.8-4.6 GHz processor and 8 GB RAM, and Python 2.7 is used as the programming language.

\subsubsection{Path planning on the 2D map}
The size of the local map is the main influencing factor of the actual flight trajectory length and computing time of each replanning step. In addition, we apply the JPS algorithm twice on two maps of different sizes and resolutions and splice the two paths into a whole. The $Map_c$ size is also a key to balancing the time cost and the path length.
As the effectiveness of our proposed method should be verified and analyzed, two rounds of numerical simulations are designed and conducted. 

\begin{figure}[]
\centering
\subfigure[]{
\includegraphics[width=0.47\textwidth]{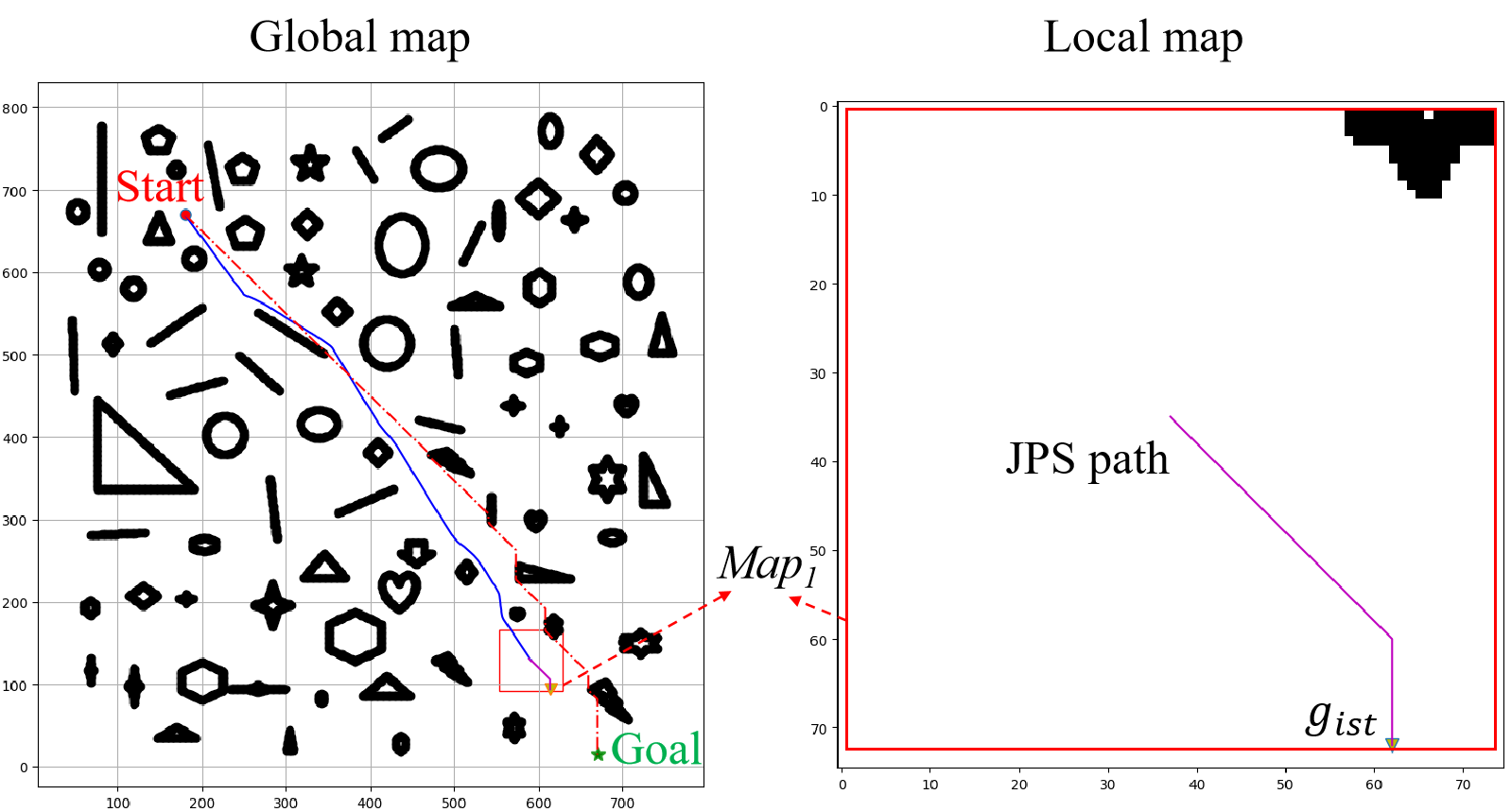}}
\hfill
\centering
\subfigure[]{
\includegraphics[width=0.47\textwidth]{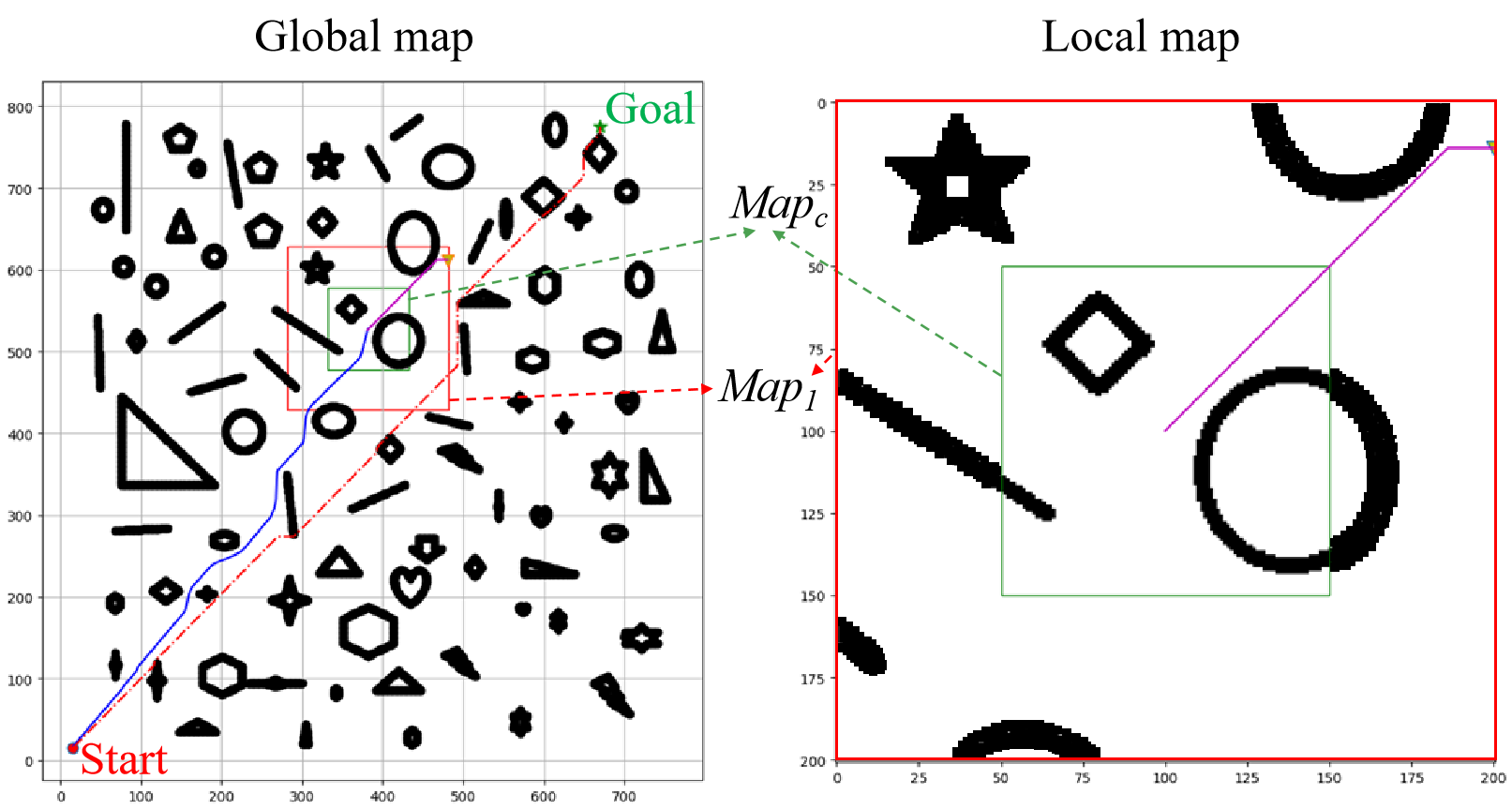}}
\caption{Visualized result during the numerical simulation. (a): only the sliding local map is used, with the map size of 75 m*75 m, (b): the double layer map is used, the sizes of $Map_1$ and $Map_c$ are 200*200 and 100*100 (unit: m). The configuration for the obstacle inflation and map downsampling is the same with Fig. \ref{fig3}. The blue line indicates the real trajectory of the drone, the red dash line indicates the global JPS path, and the purple line is the JPS path on the local map.}
\label{fig801}
\end{figure}

The first round tests the influence of the local map size, the second-round tests the effect of the $Map_c$ size (see Fig. \ref{fig3}). A large-scale 2D map is used in the numerical tests, as shown in Fig. \ref{fig801}. The map size is 800 m*800 m, and the local map size is tested with three configurations (unit: m): 75*75, 200*200, and 400*400. We assume the local map center moves along the local map path and can only move one meter (including the diagonal move) in one step. The test is conducted with 10 combinations of the randomly assigned start point and goal point, whose straight-line distance is greater than 500 m. For each local map size, the 10 combinations are identical.

For the first round, the average time cost and real trajectory length are compared with that of the planning on the entire map, as shown in TABLE \ref{tab_1}. $Len_1$ represents the average trajectory length, while $Len_2$ denotes the average global JPS path length. $T_{c1}$ is the average total computing time of each replanning step with the local map, and $T_{c2}$ is that of the global planning. TABLE \ref{tab_1} shows that $Len_1$ does not increase substantially compared to $Len_2$, while the time cost is saved considerably compared to the global planning time. We use green color to highlight the data corresponding to our proposed method, demonstrating the superiority. $Len_1$ increases by only $2.2\%$ and $T_{c1}$ decreases by $96.6\%$ compared to $Len_{2}$ and $T_{c2}$, respectively, when the $Map_1$ size is 200 m*200 m. It is the most time-efficient map size among the three tested sizes.

For the second round, the size of $Map_1$ is fixed at 200 m*200 m and the size of $Map_c$ has three alternatives (unit: m): 70*70, 100*100, and 120*120. In TABLE \ref{tab_2}, the average total computing time $T_{c3}$ and trajectory length $Len_3$ are compared between the tests that do and do not use the stitched map. When the size of $Map_c$ is 100 m*100 m, the real trajectory length $Len_3$ increases by only $0.31\%$, while the time cost $T_{c3}$ is reduced by $53.4\%$ compared to $Len_{1}$ and $T_{c1}$, respectively. Thus, the effectiveness of planning on the multi-resolution hybrid map is well validated.

\begin{table}[h]
\caption{Test results of different $Map_1$ sizes}
\label{tab_1}
\begin{center}
\begin{tabular}{|c|c|c|c|c|}
\hline
$Map_1$ size (m) & $Len_1$ (m)  & $T_{c1}$ (s) &$Len_2$ (m)& $T_{c2}$ (s)\\
\hline
75*75 & \textcolor{ForestGreen}{1021.962} &\textcolor{ForestGreen}{0.036} & 980.439 & 3.454 \\
\hline
{\textbf{200*200}} & \textcolor{ForestGreen}{\textbf{1001.783}} &\textcolor{ForestGreen}{\textbf{0.118}} &{\textbf{980.439}} & {\textbf{3.454}}
\\
\hline
400*400 &\textcolor{ForestGreen}{997.486} &\textcolor{ForestGreen}{0.284} &980.439 & 3.454\\
\hline
\end{tabular}
\end{center}
\end{table}

\begin{table}[h]
\caption{Test results of different $Map_c$ sizes}
\label{tab_2}
\begin{center}
\begin{tabular}{|c|c|c|c|c|}
\hline
$Map_c$ size (m) & $Len_3$ (m)  & $T_{c3}$ (s) &$Len_1$ (m)& $T_{c1}$ (s)\\
\hline
70*70 & \textcolor{ForestGreen}{1110.374} &\textcolor{ForestGreen}{0.040} & 1001.783 & 0.118 \\
\hline
\textbf{100*100} & \textcolor{ForestGreen}{\textbf{1004.848}} &\textcolor{ForestGreen}{\textbf{0.055}} & \textbf{1001.783} & \textbf{0.118}\\
\hline
120*120 &\textcolor{ForestGreen}{1002.917} &\textcolor{ForestGreen}{0.082}& 1001.783 & 0.118\\
\hline
\end{tabular}
\end{center}
\end{table}

\subsubsection{Shorter 3D path searching}

\begin{figure}[]
\centering
\subfigure[]{
\includegraphics[width=0.46\textwidth]{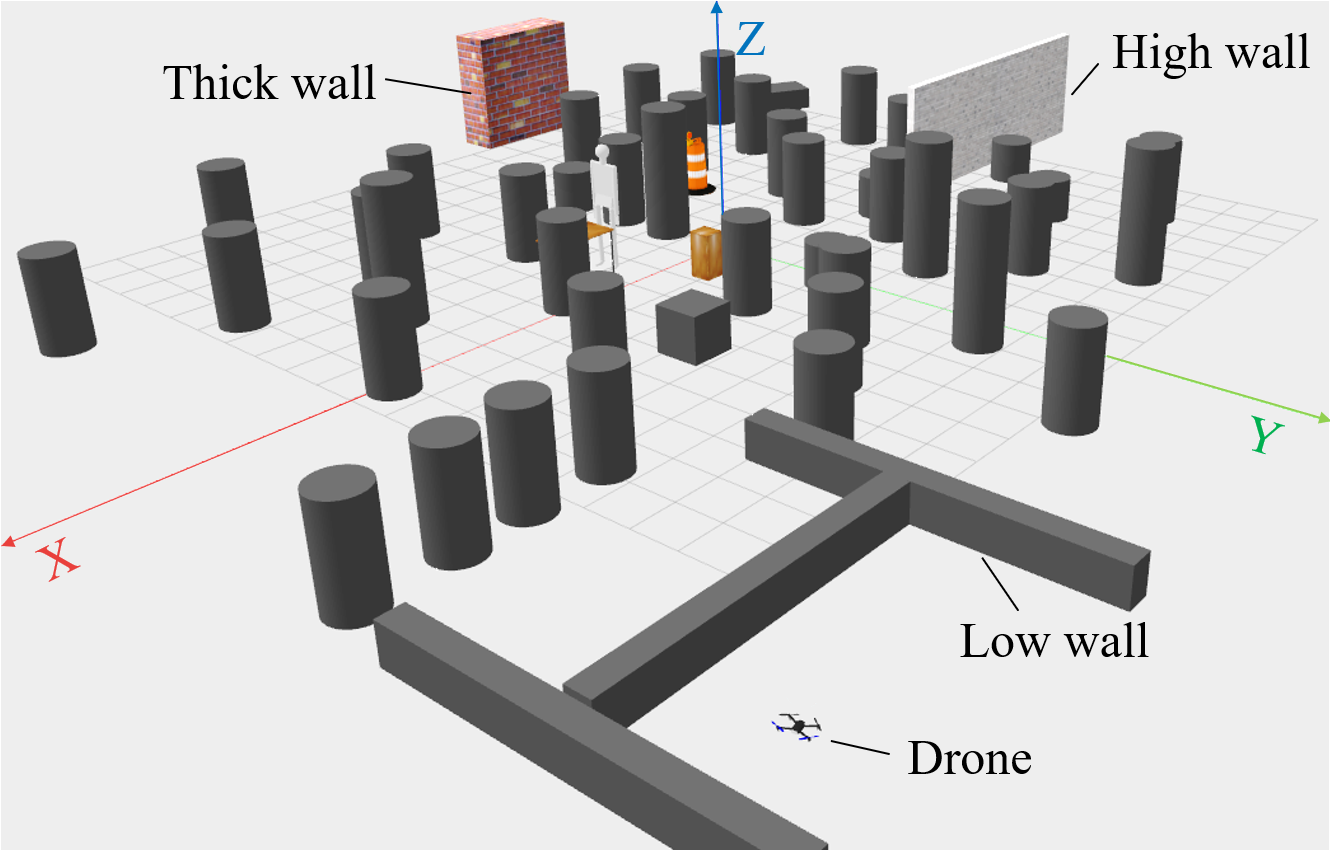}}
\hfill
\centering
\subfigure[]{
\includegraphics[width=0.41\textwidth]{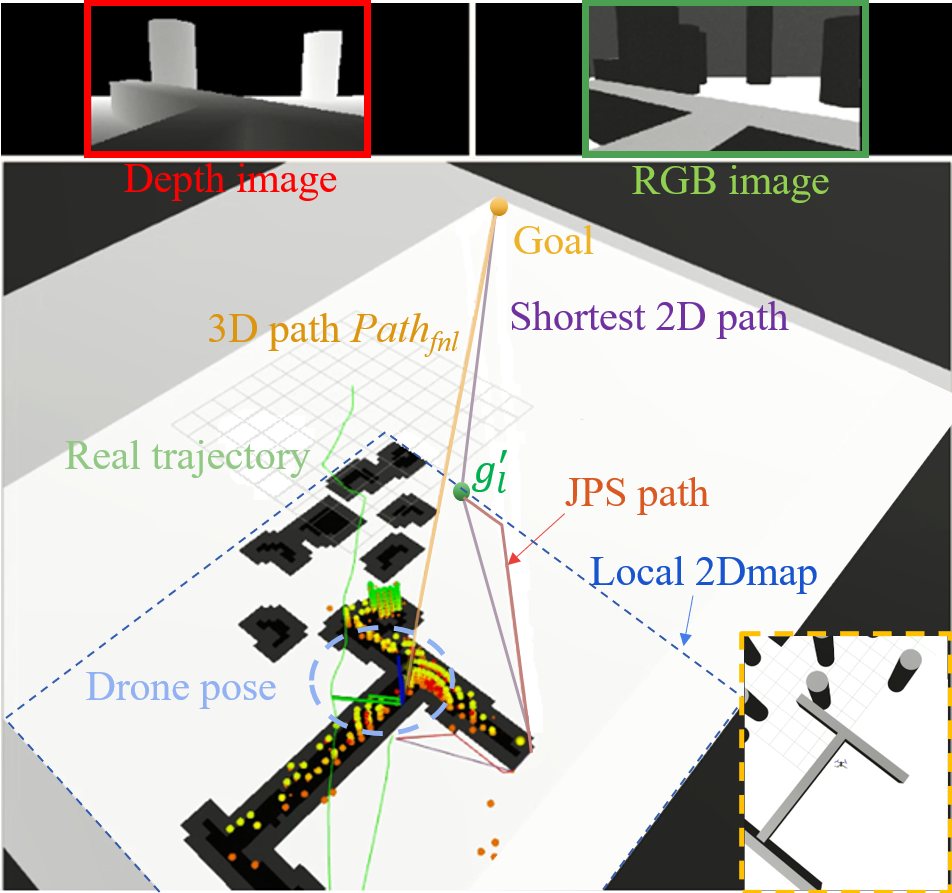}}
\caption{(a): The simulation world. (b): The visualized data in RVIZ. The Gazebo window when the flight test is ongoing is shown at the lower right corner. The colorful dots is the point cloud of the 3D local map. The black blocks on the ground plane in RVIZ stand for the obstacles, the white part stands for the free or unknown area.}
\label{fig802}
\end{figure}

To study the path length shortened by the 3D path search and corresponding extra time cost, the flight data in the Gazebo/ROS simulation environment is analyzed. Gazebo is a simulation software that provides a physical simulation environment similar to the real world. Compared to our experimental hardware platform, all the simulation configurations are set up as the same or similar to ensure the credibility of the simulation and the analysis conclusion. 
The Gazebo simulation world and the visualized data are shown in Fig. \ref{fig802}. The obstacle feature size in the simulation is from 0.5 m to 6 m, which is similar to that of most real scenes. Three local 3D map bottom sizes are used (unit: m): 12*12, 20*20, and 30*30, and the map height is fixed at 6 m with the map resolution of 0.2 m. The drone is at the center of the local map. We also conduct 5 flight tests with different combinations of starting and goal points for each local map size. For each flight test, an additional flight test without the 3D path search procedure is used as the control group to compare the final trajectory length. During the flight simulation, the time cost of the 3D path search and the path length is recorded for statistical analysis. The results can be found in TABLE \ref{tab_3}. $\eta_{2D}$ indicates the mean of shortening percentage of the 3D path compared to the 2D path. $T_{3D}$ is the average time cost for the 3D path search (Algorithm 3). $Len_{3D}$ and $Len_{2D}$ are the actual trajectory average lengths for the flight tests with and without Algorithm 3, respectively. 

\begin{table}[h]
\caption{Flight test results for the 3D path planning}
\vspace{-0.5cm}
\label{tab_3}
\begin{center}
\resizebox{9 cm}{10mm}{
\begin{tabular}{|c|c|c|c|c|}
\hline
Map size (m) & $T_{3D}$ (s)  & $\eta_{2D}$ ($\%$)  &$Len_{3D}$ (m)& $Len_{2D}$ (m)\\
\hline
12*12 &\textcolor{ForestGreen}{0.011} & \textcolor{ForestGreen}{38.6} & \textcolor{ForestGreen}{39.274} & 47.263\\
\hline
\textbf{20*20} &\textcolor{ForestGreen}{\textbf{0.032}}  &\textcolor{ForestGreen}{\textbf{35.7}} &\textcolor{ForestGreen}{\textbf{32.635}} & \textbf{44.582}\\
\hline
30*30  &\textcolor{ForestGreen}{0.059}  & \textcolor{ForestGreen}{35.0} & \textcolor{ForestGreen}{30.843} & 42.341\\
\hline
\end{tabular}}
\end{center}
\end{table}

We can see that $\eta_{2D}$ and $Len_{3D}$ decrease as the map size increases. When the map size is small, the points in the 3D map reduces. Seemingly, the algorithm is more likely to find a shorter 3D path. However, the 3D path of a small local map is more likely to collide with the newly appearing obstacles in the updated local map. Therefore, the drone has a high probability of taking a detour while flying along the 3D path, and the actual trajectory length is longer than when we use a large local map. When the local map size is settled as 20 m*20 m, the average outer loop frequency of the path planning is greater than 15 Hz, and the actual trajectory is smooth and natural. We use this map size in the following flight tests.

\subsubsection{The improvements in optimization formula}
After several hardware flight tests with our proposed framework, we record all the required data for solving the optimization problem, including $p_n$, $v_n$, and $w_{pn}$, at each step (over $5.3\times 10^{4}$ steps in total). To validate the improvements of the optimization formula, the collected data is input to the original optimization formula and the improved one in this paper for comparison. In addition, the optimization solving performance under different maximum iterating numbers is studied. The average time cost and overall success rate are counted for quantitative comparison. In TABLE \ref{tab_opti}, $R_{og}$ and $T_{og}$ are the solution success rate and the average time cost of the original optimization formula, respectively, and $R_{im}$ and $T_{im}$ are for the improved version. We can see that the success rate and time cost are greatly improved. When the maximum step is more than 20, the success rate improvement is minor. Because $99.83 \%$ is a satisfactory success rate, we set the maximum step number as 20 to reduce the time cost. The motion optimization problem-solving time decreases by $39.33\%$ compared to that of the original formula.

\begin{table}[h]
\caption{Test results for the improvements of optimization formula}
\label{tab_opti}
\begin{center}
\begin{tabular}{|c|c|c|c|c|}
\hline
Max steps & $R_{og}$ ($\%$)  & $R_{im}$ ($\%$)  &$T_{og}$ (ms)& $T_{im}$ (ms) \\
\hline
5 & $41.51$ &\textcolor{ForestGreen}{89.12 }&3.24  & \textcolor{ForestGreen}{2.12} \\
\hline
10 & $61.14$ &\textcolor{ForestGreen}{95.49 }&4.87  &\textcolor{ForestGreen}{2.94} \\
\hline
\textbf{20}  &\textbf{$76.87$} &\textcolor{ForestGreen}{\textbf{99.83}}&6.56  &\textcolor{ForestGreen}{\textbf{3.98}} \\
\hline
40 &$83.68$ &\textcolor{ForestGreen}{99.96} &9.45  &\textcolor{ForestGreen}{5.23} \\
\hline
80 &$92.15 $ &\textcolor{ForestGreen}{100.00} &14.78  &\textcolor{ForestGreen}{5.61} \\
\hline
\end{tabular}
\end{center}
\end{table}

\subsection{Simulated flight tests with real-time planning}

Compared to the counterpart that follows the original JPS path directly on the 2D map, our proposed framework is proved to shorten the actual trajectory substantially with limited additional time cost. Another test is required to compare the final trajectory length with the 3D global shortest path to further validate the framework's performance. However, the globally optimal path can only be obtained after the map is entirely constructed, so we cannot obtain it while exploring the environment. Because the 3D map of the environment is represented by the point cloud, no graph-searching based algorithm can be applied to obtain the 3D optimal path length. We adopt the asymptotically optimal method RRT* to generate the globally optimal path using a large amount of offline iteration computing. For comparison, the same simulation configuration with the former flight tests is used in this test. We first fly the drone manually with the mapping kit to build up the globally 3D map for the simulation world. Then, the RRT* method is applied 5 times on the map with each group of starting and goal points. The shortest path of the 5 runs is considered the global optimal. The initial parameters of RRT* are generated randomly so that the repeating can avoid the local optimum. The iteration terminates when the relative error of the path length is smaller than $10^{-3}$ in the last 10 iterations. Table \ref{table_1} describes the parameter settings of the framework in the simulation test. ``pcl'' is short for the point cloud. 

\begin{figure}[]
\centering
\subfigure[]{
\includegraphics[width=0.48\textwidth, height=4.8cm]{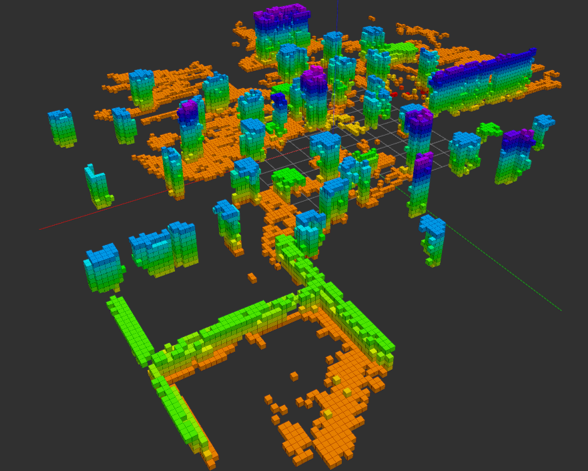}}
\hfill
\centering
\subfigure[]{
\includegraphics[width=0.47\textwidth]{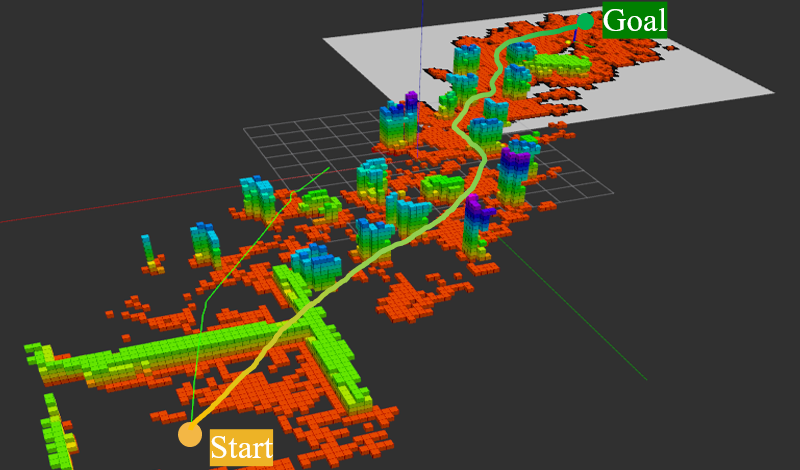}}
\hfill
\centering
\subfigure[]{
\includegraphics[width=0.49\textwidth]{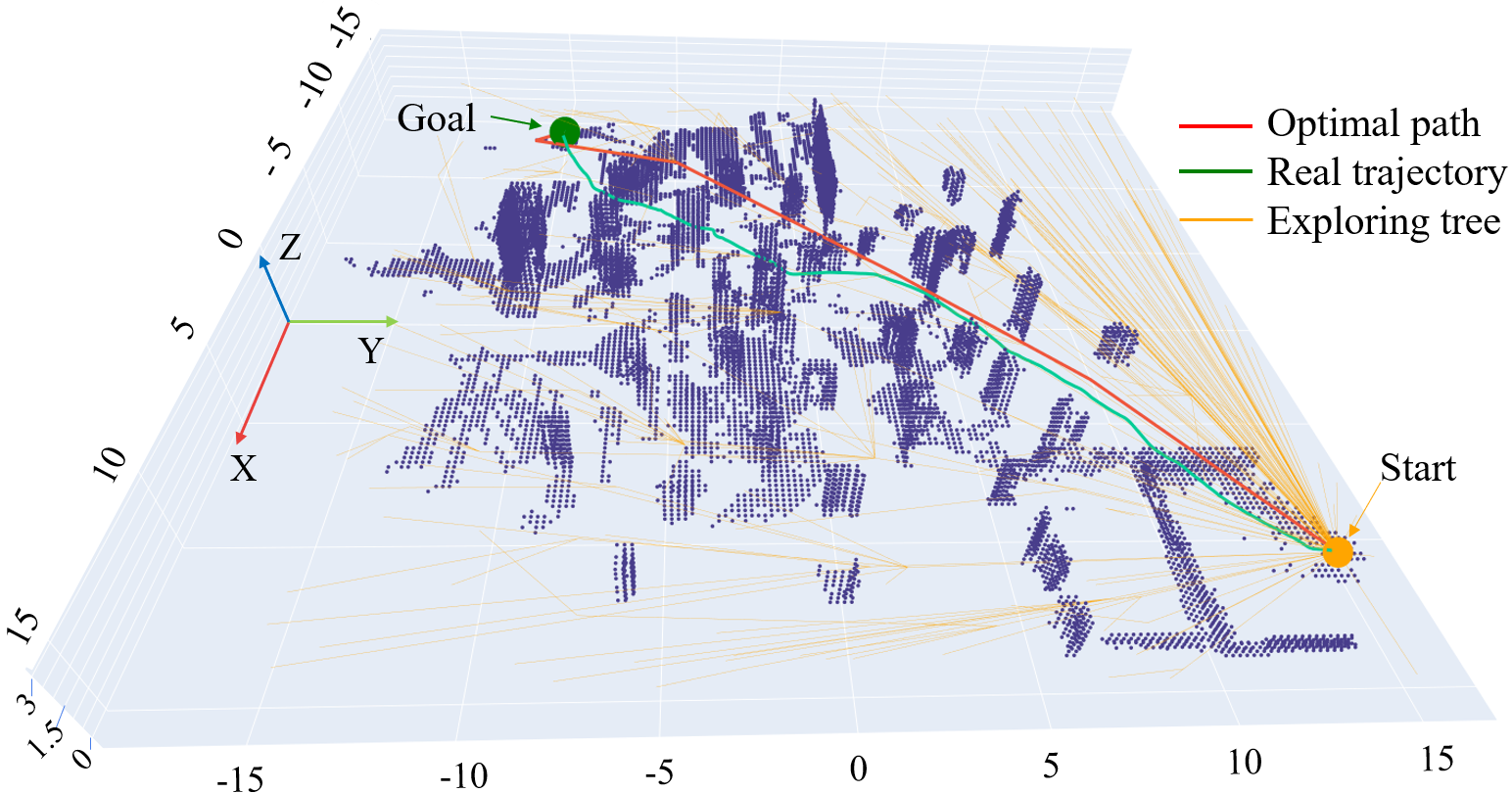}}
\caption{(a): The entire 3D occupancy grid map. (b): The map after one flight test using our proposed framework. The gradient color curve represents the final trajectory. (c): The comparison of the globally optimal path found by RRT* and the actual trajectory from the flight test data. The 3D map is shown in the form of a point cloud. (b) and (c) are for the same flight test from different views. }
\label{fig803}
\end{figure}

All the length results of the 10 flight tests are shown in TABLE \ref{tab_4}. The mean of the actual trajectory length $Len_{3D}$ increases by approximately $12.8 \%$ compared to the mean of the globally optimal path length $Len_{opt}$. $t_{mp}$ is the average step time cost of MP in this flight. $t_{rrt*}$ is the total time cost for RRT* to find the shortest path in the 5 runs. $t_{rrt*}$ only include the time cost of the runs before the shortest path is found. Fig. \ref{fig803} illustrates the detailed visualized data of the flight test corresponding to second test in TABLE \ref{tab_4} (the bold part).

Fig. \ref{fig803}(a) demonstrates the entire map of the simulation world in Fig. \ref{fig802}. The coordinates of the starting and goal points of the flight test in Fig. \ref{fig803}(b) are $(11.4,14.4,0.8)$ and $(-12.0,-10.0,1.2)$, respectively. The global optimal path length for this flight is 33.252 m, and the actual flight trajectory length is 36.057 m. We see that the real trajectory has obvious differences with the globally optimal path. Nevertheless, the path length gap is not large, which is similar to the numerical test results on the 2D map.

The trajectory length comparison between our proposed framework and the state-of-the-art algorithms is shown in TABLE \ref{tab_5}. \cite{chen2020computationally} is our former work for the trajectory planner. $\eta_{3D}$ indicates the mean of the 3D path length percentage increase compared to the optimal 3D path. Except for the $\eta_{3D}$ of \cite{chen2020computationally} is evaluated from the flight test data in the same simulation world, the other data is obtained from the experimental results in the references. 
Our proposed framework can fly the shortest trajectory compared to the works listed in TABLE \ref{tab_5}.

\begin{table}[h]
\caption{Parameters for the framework}
\label{table_1}
\begin{center}
\begin{tabular}{|c|c|c|c|}
\hline
Parameter& Value &Parameter & Value\\
\hline
$ l_{ms}$ & 20 m & $h_{ms}$ & 6 m\\
\hline
$\alpha_{res}$ & $10^{\circ}$ &$i,j$ &100\\
\hline
$m,n$ &50  &$k$ &3\\
\hline
$h$ & 2 &$r_{safe}$ &0.5 m\\
\hline
$\overline{p_{n}w_{pn}}$& 0.3 m & $n_{use}$ & 70\\
\hline
voxel size & $0.2$ m & pcl frequency & 30 Hz\\
\hline
depth resolution & 640*360 & $\kappa_{1},\ \kappa_{2}$ & 4.2, 1.5\\
\hline
$\eta_{1},\eta_{2}$ &40, 10 &$r_{dec}$ & 3 m\\
\hline
\end{tabular}
\end{center}
\end{table}

\begin{table}[h]
\caption{3D path length comparison with the global optimal}
\label{tab_4}
\begin{center}
\begin{tabular}{|c|c|c|c|c|c|}
\hline
$Len_{3D}(m)$ & $Len_{opt}(m)$ &$t_{mp} (s)$ & $t_{rrt*}(s)$\\
\hline
\textcolor{ForestGreen}{41.385} &36.761 &\textcolor{ForestGreen}{0.073} &6.294 \\
\hline
\textcolor{ForestGreen}{\textbf{36.057}} &\textbf{33.252} &\textcolor{ForestGreen}{\textbf{0.067}}&\textbf{6.091}\\
\hline
\textcolor{ForestGreen}{32.249} &26.825 &\textcolor{ForestGreen}{0.068} &4.505\\
\hline
\textcolor{ForestGreen}{30.674} &27.692 &\textcolor{ForestGreen}{0.072} &3.974\\
\hline
\textcolor{ForestGreen}{38.668} &32.707 &\textcolor{ForestGreen}{0.071} &5.340\\
\hline
\textcolor{ForestGreen}{34.916} &31.832 &\textcolor{ForestGreen}{0.069} &6.110\\
\hline
\textcolor{ForestGreen}{38.269} &33.403 &\textcolor{ForestGreen}{0.074} &4.904\\
\hline
\textcolor{ForestGreen}{39.475} &36.412 &\textcolor{ForestGreen}{0.073} &3.841\\
\hline
\textcolor{ForestGreen}{33.785} &31.353 &\textcolor{ForestGreen}{0.068} &5.916\\
\hline
\textcolor{ForestGreen}{40.879} &36.044 &\textcolor{ForestGreen}{0.070} &6.554\\
\hline
\end{tabular}
\end{center}
\end{table}

\begin{table}[h]
\caption{3D path length comparison with the state-of-the-art algorithms}
\label{tab_5}
\begin{center}
\begin{tabular}{|c|c|c|c|c|c|}
\hline
work&$\eta_{3D}$ & work&$\eta_{3D}$& work&$\eta_{3D}$\\
\hline
\textcolor{ForestGreen}{\textbf{Ours}} & \textcolor{ForestGreen}{$\textbf{12.8\%}$} &\cite{tordesillas2019real} & $15.6\%$ &\cite{oleynikova2018safe} & $29.7\%$\\
\hline
\cite{bircher2016receding}& $49.6\%$ &\cite{chen2020computationally} & $29.5\%$& \cite{zhou2019robust}& $34.5\%$\\
\hline
\end{tabular}
\end{center}
\end{table}

\subsection{Hardware flight tests}

The video for the hardware test has been uploaded online. In the test environment, static and dynamic obstacles are present to validate the fast reaction of the PCP.

\subsubsection{Introduction of hardware platform}

We conduct the hardware tests on a self-assembled quadrotor. The Intel RealSense depth camera D435i is installed under the frame as the only perception sensor. The drone frame is QAV250, with the diagonal length 25 cm. The Pixracer autopilot with a V1.10.1 firmware version is adopted as the underlying flight controller. A LattePanda Alpha 800S with an Intel Core M3-8100Y, dual-core, 1.1-3.4 GHz processor is installed as the onboard computer, where all the following timing breakdowns are measured. For the hardware test, we fly the drone in multiple scenarios with our self-developed visual-inertial odometry (VIO) kit\footnote{https://github.com/HKPolyU-UAV/FLVIS} to demonstrate the practicality of our proposed framework. The point cloud filter, the VIO kit, and the mapping kit are executed by C++ code, while the other modules are run by the Python scripts. All the parameters used in the hardware test are identical to the simulation flight tests, as shown in TABLE \ref{table_1}. The yaw angle of the drone is controlled to keep the camera always heading toward the local goal $g'_{l}$.

\subsubsection{Indoor tests}

\begin{figure}[]
      \centering
      \includegraphics[width=0.49\textwidth]{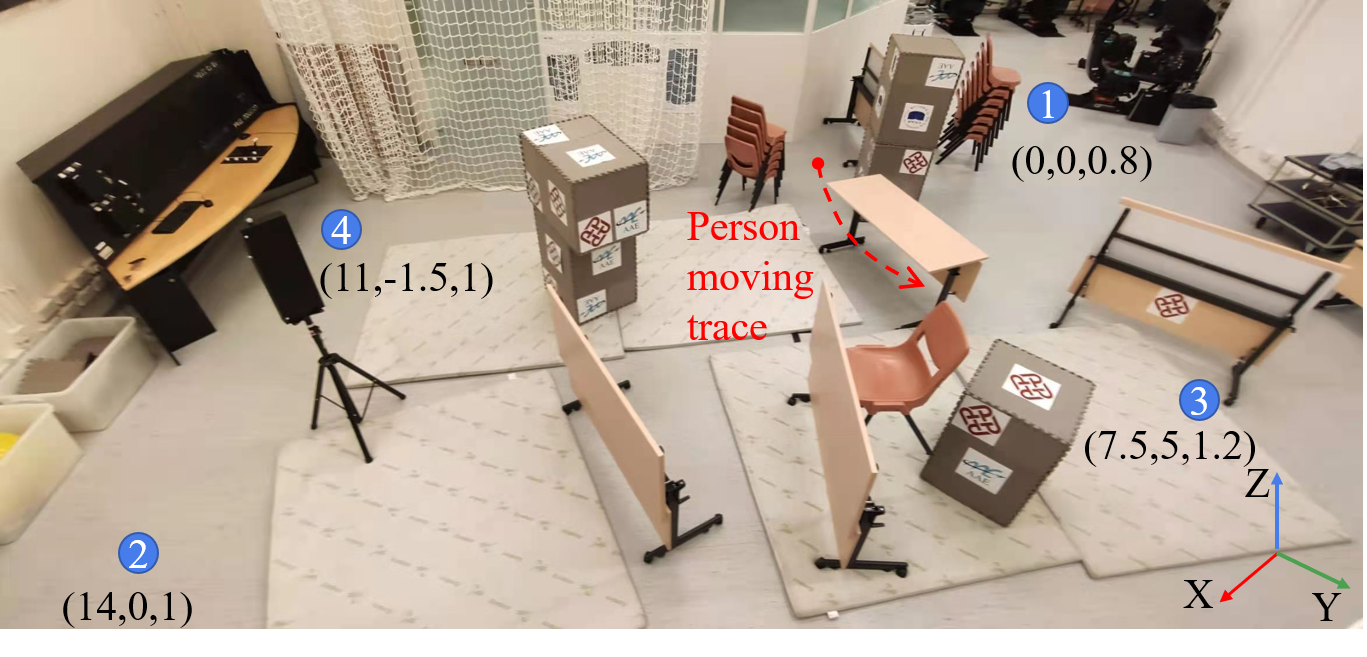}
      \caption{Indoor environment for the hardware flight tests. In the video, a flight test is first conducted with static obstacles, and then the table closest to trace of the person moving is removed for the test with an intruding person.}
      \label{fig8}
   \end{figure}

Fig. \ref{fig8} is our indoor flight test environment. First, the drone takes off from point 1 and flies through the cluttered static obstacles (shown in the picture), following the sequence 2-3-4. The environment is unknown to the drone before it takes off. We can see from the video that the drone can avoid the obstacles agilely. Meanwhile, the drone posture is stable and the final flight trajectory is smooth. After the drone reaches point 3 and starts flying toward point 4, a person who hides behind the boxes suddenly appears closely in front of the drone. In the video, the drone quickly maneuvers after the person appears, and the avoidance is successful. The map is updated afterward, and the drone continues to follow the path from the MP.

\begin{figure}[]
      \centering
      \includegraphics[width=0.49\textwidth]{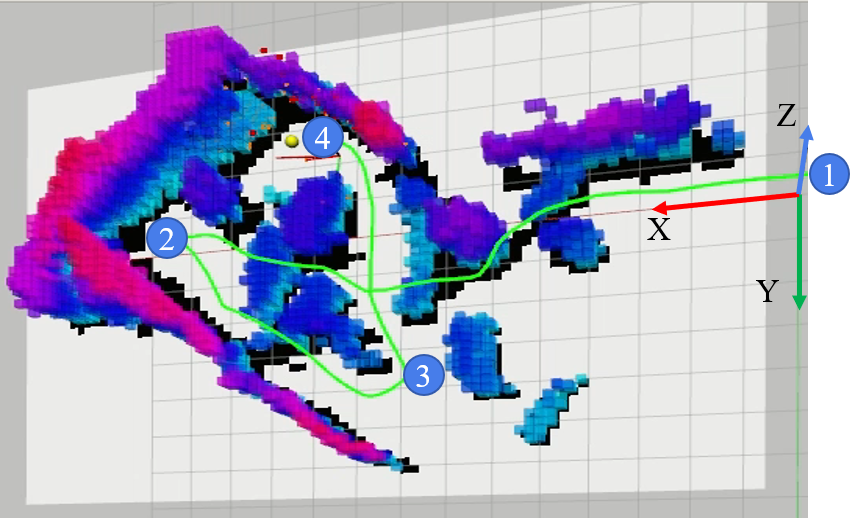}
      \caption{Explored map and the drone trajectory after the hardware flight test in the static environment.}
      \label{fig81}
   \end{figure}

\begin{figure}[]
\centering
\subfigure[]{
\includegraphics[width=0.48\textwidth]{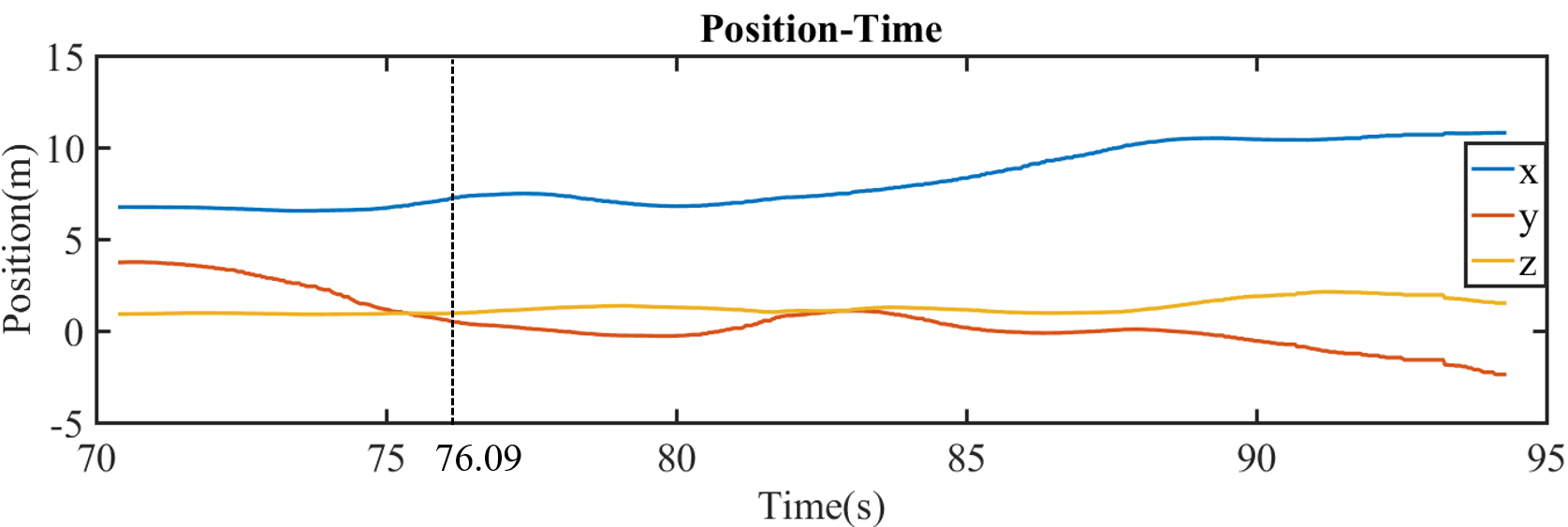}}
\hfill
\centering
\subfigure[]{
\includegraphics[width=0.48\textwidth]{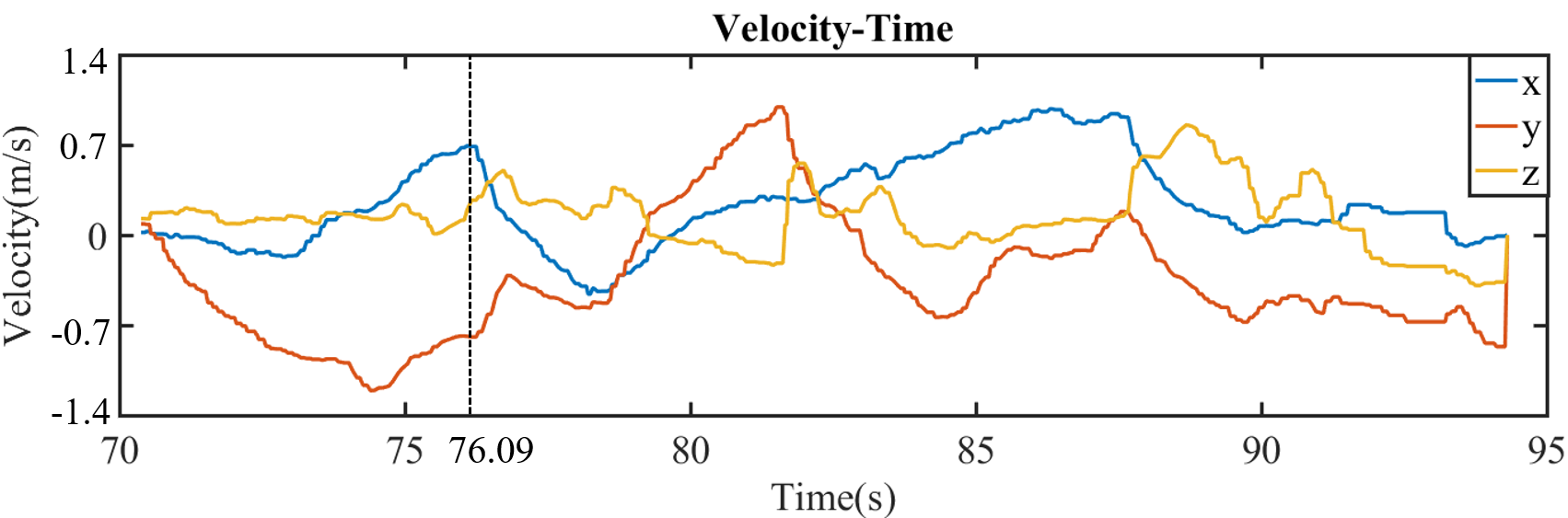}}
\hfill
\centering
\subfigure[]{
\includegraphics[width=0.48\textwidth]{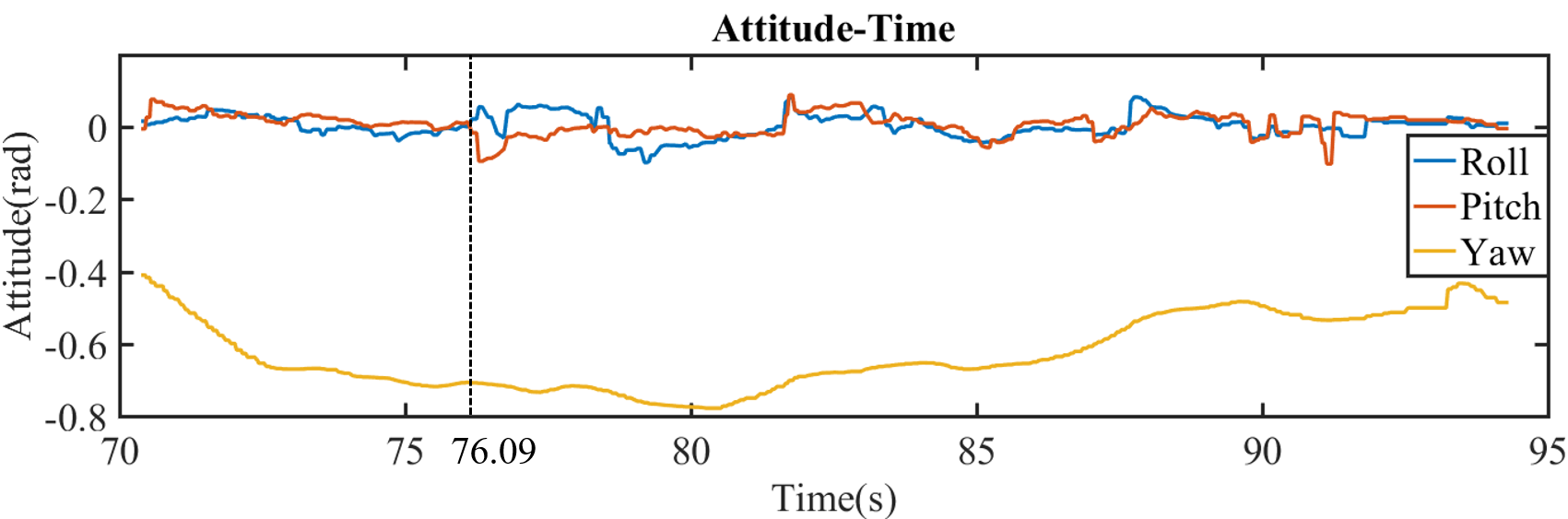}}
\caption{(a)-(c): Curves of the three-axis coordinate positions, flight velocities, and attitude angles. The framework begins to work at time 0 s, and the data shown in the figures start at 70.38 s and end at 94.31 s, corresponding to the flight from point 3 to point 4. The moving person enters the depth camera's FOV at 76.09 s (marked with the vertical dashed lines).}
\vspace{-0.4cm}
\label{fig9}
\end{figure}

The constructed voxel map and flight trajectory after this flight are shown in Fig. \ref{fig81}. The position, attitude, and velocity curves of the drone can be found in Fig. \ref{fig9}. The attitude angles react immediately to the appearance of the person, and then the velocity changes considerably. Sequentially, the drone decelerates and flies to the left side to pass the person. The maximum speed is $1.23$ m/s at $74.40$ s. The drone does not approach any obstacles before this time, so the speed continues to increase. Because $\|\overrightarrow{p_{n}w_{pn}}\|_{2}$ is assigned to be always greater than $\|\overrightarrow{p_{n}p_{n+1}}\|_{2}$, in the motion optimization problem (7), minimizing the cost $\eta_{1} \| \overrightarrow{w_{pn}p_{n+1}}\|_{2}$ leads to positive acceleration. The drone will decelerate when the obstacles are near it, because of the acceleration cost and the endpoint cost in the objective function.

\subsubsection{Outdoor tests}
\begin{figure}[]
\centering
\subfigure[]{
\includegraphics[width=0.233\textwidth]{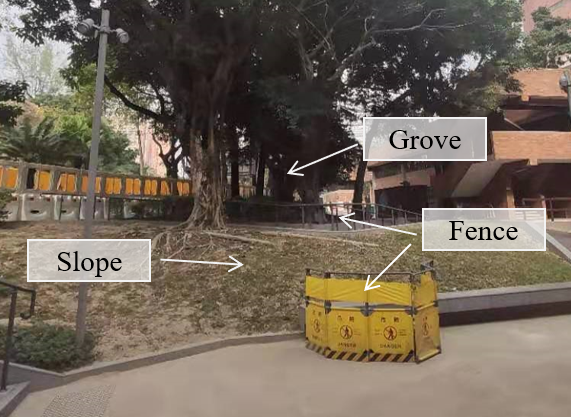}}
\hfill
\centering
\subfigure[]{
\includegraphics[width=0.233\textwidth]{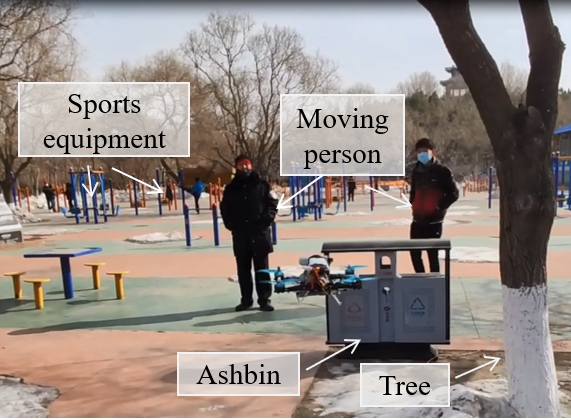}}
\caption{Two of the outdoor flight test environments. (a) locates in a campus and (b) is at the sports corner in a park.}
\label{fig812}
\end{figure}

In addition, the proposed framework is tested in diverse outdoor scenarios. Fig. \ref{fig812} shows the environments of two experiments, and the flight tests in other environments are included in the video. Fig. \ref{fig812}(a) is entirely static, with several obstacles standing on a slope and dense bushes and trees. This environment is challenging, requiring the 3D precise trajectory planning and motion control. Fig. \ref{fig812}(b) is a larger environment compared to those of the tests above. In addition to the complex static obstacles, five moving people in the field continuously interrupt the drone from the original planned path and validate its reaction performance. The video demonstrates that the drone performs agile and safe flights in various test environments, so the practicability and flight efficiency of our proposed framework can be proven.
   
\begin{figure}[]
\centering
\subfigure[]{
\includegraphics[width=0.43\textwidth]{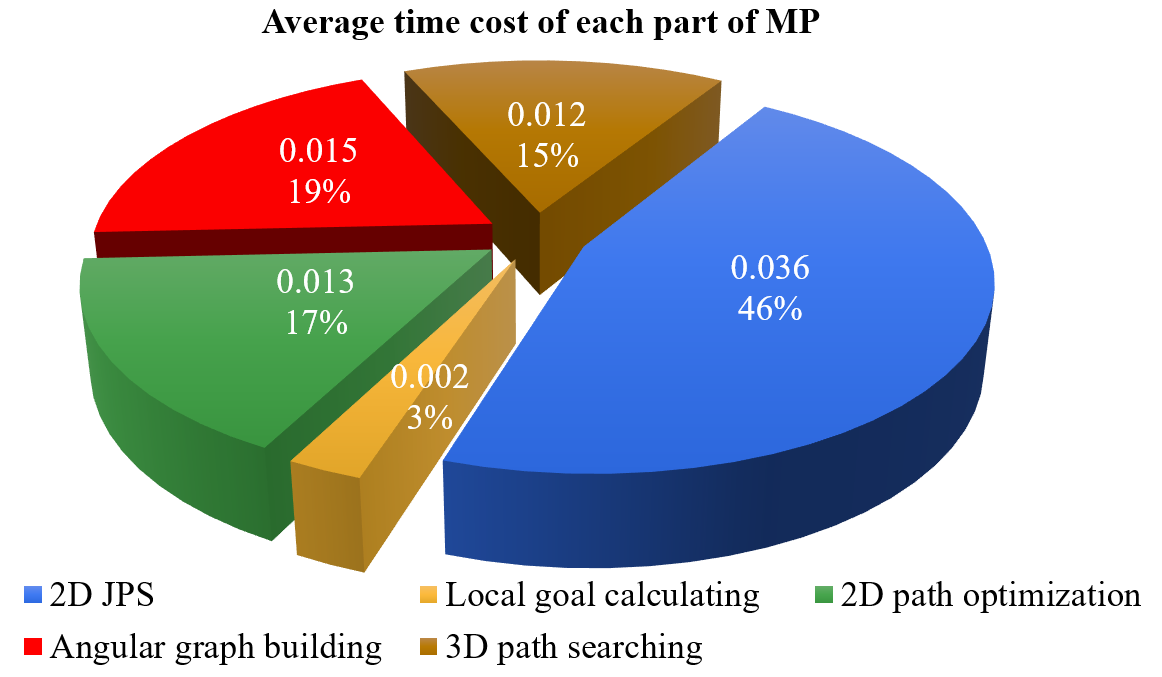}}
\hfill
\centering
\subfigure[]{
\includegraphics[width=0.49\textwidth]{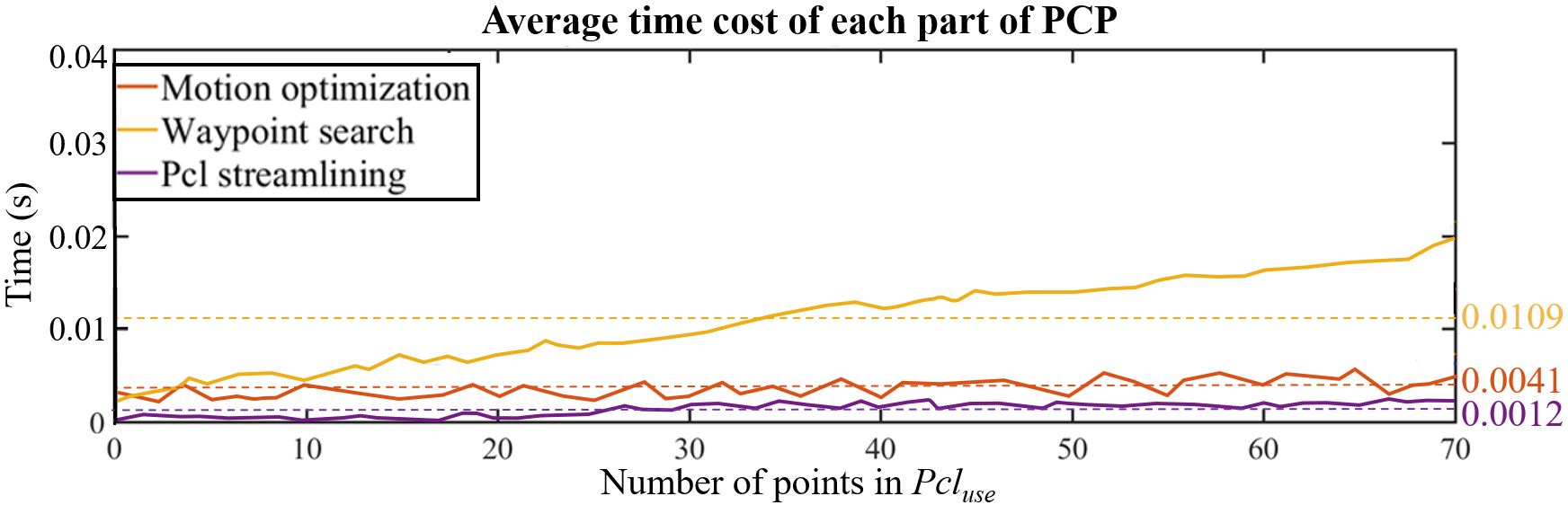}}
\caption{(a): Average time cost and the proportion of each submodule of MP. (b): The time cost versus $Pcl_{use}$ size curves of each part of the PCP. The average time cost is marked with a dashed line, and the values are on the right side.}
\label{fig911}
\end{figure}

Finally, the average time cost of each part of the MP and PCP for the hardware tests is counted and analyzed in Fig. \ref{fig911} to show the computational efficiency. In Fig. \ref{fig911}(a), the average time cost and the percentage are provided on a pie chart. In Fig. \ref{fig911}(b), the relationship between the time cost of each procedure in the PCP and the size of $Pcl_{use}$ is illustrated, with the average time cost shown on the right side. For the MP, most of the time cost ($63\%$) is used for path planning and optimization on the 2D map. The path search with the 3D point cloud is computationally inexpensive. The average total time cost of each MP step is 0.078 s, and the loop frequency is approximately 12 Hz. These results are slower than the offline test results because the computing resource is occupied by the other part of the framework (VIO, mapping kit, point cloud filter, and the PCP). For the PCP, only the time cost of the waypoint searching is relevant to the $Pcl_{use}$ size because the number of the points determines the collision check's circling number. The average time cost of the PCP step is 16.2 ms, of which searching for $w_{pn}$ is the most time-consuming part.

Moreover, the time cost is compared with those of the state-of-the-art algorithms in TABLE \ref{table_2}. Because our proposed method is composed of two planners running asynchronously, no single value represents the framework execution time. Thus, the average step time costs of the MP and PCP are listed for the comparison. Notably, the time costs of the related works are measured on different hardware platforms with different program code types. ``MSCF'' is the abbreviation for the maximum single-core frequency of the hardware platform processor. Although TABLE \ref{table_2} cannot be used for the absolute performance comparison, the trajectory replanning time cost of our proposed method (PCP) is believed promisingly to be better than those of the state-of-the-art algorithms.

\begin{table}[thpb]
\setlength{\belowcaptionskip}{0.5cm}
\caption{COMPARISON WITH STATE-OF-THE-ART ALGORITHMS.}
\label{table_2}
\begin{center}
\begin{tabular}{|c|c|c|}
\hline
Works & Time cost (ms) & MSCF (GHz)\\
\hline
MP & 78 & 3.4 \\
\hline
\textcolor{ForestGreen}{\textbf{PCP}} & \textcolor{ForestGreen}{\textbf{16.2}} & \textcolor{ForestGreen}{\textbf{3.4}}\\
\hline
\cite{zhou2019robust} & $>$100 & 3.0\\
\hline
\cite{liu2016high} &$>$160 & 3.4\\
\hline
\cite{burri2015real} & $>$40 & N/A \\
\hline\cite{chen2020computationally} & 19 & 4.6\\
\hline
\cite{bircher2016receding}& 199 & 3.1 \\
\hline\cite{gao2018online} &106 & 3.0\\
\hline
\end{tabular}
\end{center}
\vspace{-0.4cm}
\end{table}

\section{Conclusion and future work}
In this paper, a framework of trajectory planning for UAVs with two parallel planners is introduced. The map planner tries to find the shortest possible path in limited computational time. The point cloud planner takes effect when the point cloud near the drone differs from the 3D map to ensure safety. It reacts much faster than the path planning on the map. The test results verify that the techniques proposed in this paper can reduce the computing time cost, with the performance basically unchanged or even improved compared to that of the original method \cite{chen2020computationally}. The real-time flight trajectory length outperforms those of the state-of-the-art algorithms, and the reacting time of the PCP is also superlative. The entire framework is tested extensively in simulation and hardware experiments, demonstrating excellent rapid response capabilities and flight safety.

However, the test environments do cover all the UAV application scenarios. Moreover, the UAV flight speed in our tests is not sufficiently high. In the future, the framework will be tested in more challenging environments with a higher vehicle speed than the current study.


%



\section*{Acknowledgment}

The authors would like to thank Mr. Ching-wei Chang for his kind help in the hardware equipment debugging and Miss Yuyang Hu for her assistance in the hardware tests.

\ifCLASSOPTIONcaptionsoff
  \newpage
\fi

\bibliographystyle{ieeetr}

\bibliography{bare_jnl.bbl}

\begin{IEEEbiography}[{\includegraphics[width=1in,height=1.25in,clip,keepaspectratio]{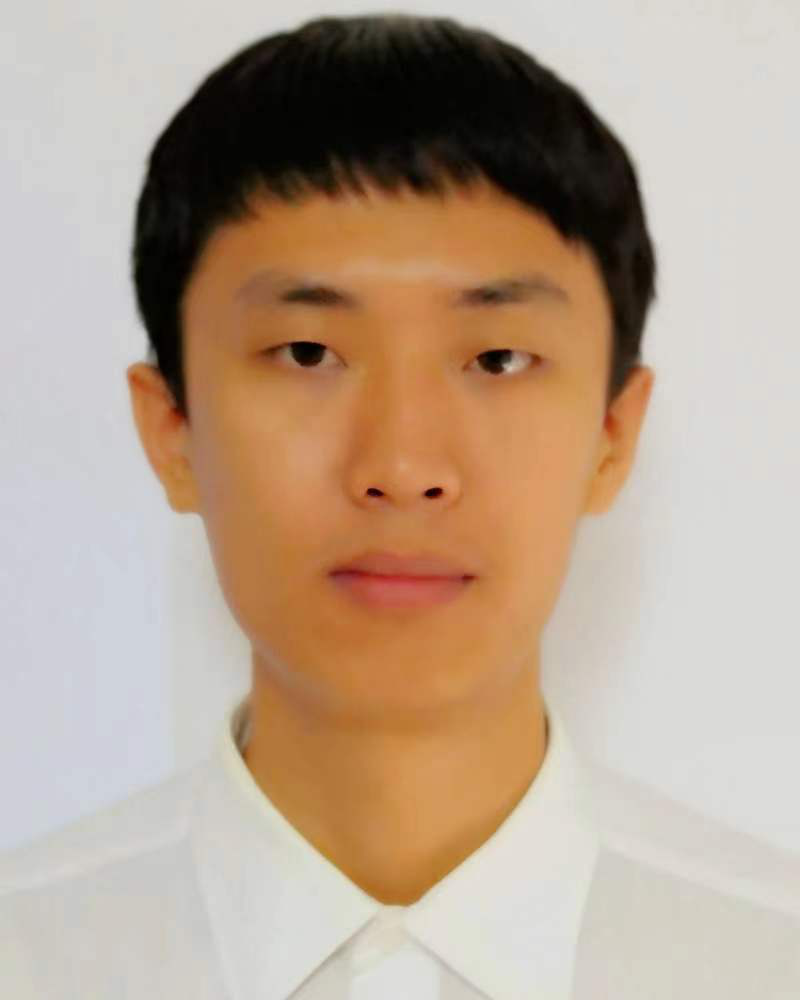}}]{Han Chen}
received his Bachelor of Engineering in 2016 from Beijing Institude of Technology, P.R.China and Master of Science from Beijing Institude of Technology in 2019. Currently, he is a PhD candidate in MAV/UAV Lab and ARC Lab, Department of Aeronautical and Aviation Engineering, The Hong Kong Polytechnic University.
\end{IEEEbiography}

\begin{IEEEbiography}[{\includegraphics[width=1in,height=1.25in,clip,keepaspectratio]{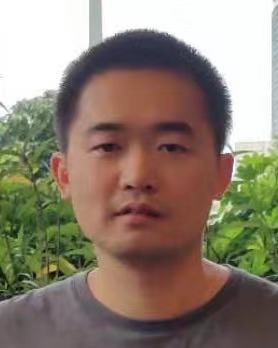}}]{Shengyang Chen}
received his Bachelor of Engineering from Northwestern Polytechnical University, P.R.China and Master of Science from University of Siegen, Germany respectively. Currently, He is a PhD candidate in MAV/UAV Lab, Department of Mechanical Engineering, The Hong Kong Polytechnic University.
\end{IEEEbiography}

\begin{IEEEbiography}[{\includegraphics[width=1in,height=1.25in,clip,keepaspectratio]{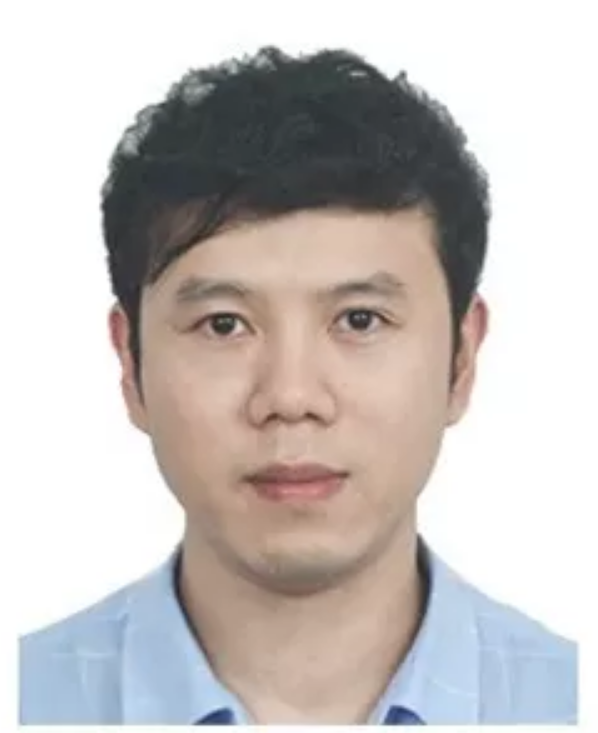}}]{Peng Lu}
obtained his BSc degree in automatic control and MSc degree in nonlinear flight control both from Northwestern Polytechnical University (NPU). He continued his journey on flight control at Delft University of Technology (TU Delft) where he received his PhD degree in 2016. After that, he shifted a bit from flight control and started to explore control for ground/construction robotics at ETH Zurich (ADRL lab) as a Postdoc researcher in 2016. Since 2020 he works in the University of Hong Kong, Department of Mechanical Engineering, as an assistant professor.
\end{IEEEbiography}


\begin{IEEEbiography}[{\includegraphics[width=1in,height=1.25in,clip,keepaspectratio]{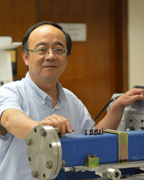}}]{Chih-Yung Wen}
received his Bachelor of Science degree from the Department of Mechanical Engineering at the National Taiwan University in 1986 and Master of Science and PhD from the Department of Aeronautics at the California Institute of Technology (Caltech), U.S.A. in 1989 and 1994 respectively. He joined the Department of Mechanical Engineering, The Hong Kong Polytechnic University in 2012, as a professor. In 2019, he became the interim head of the Department of Aeronautical and Aviation Engineering in The Hong Kong Polytechnic University. His current research interests include modeling and control of tail-sitter UAVs, visual-inertial odometry systems for UAVs and AI object detection by UAVs.
\end{IEEEbiography}




\end{document}